\documentclass[journal, twocolumn]{IEEEtran}

\usepackage[T1]{fontenc}
\usepackage{url}
\usepackage[hidelinks]{hyperref}
\usepackage[utf8]{inputenc}
\usepackage[small]{caption}
\usepackage{amsmath}
\usepackage{amsthm}
\usepackage{booktabs}
\usepackage{algorithm}
\usepackage{algorithmic}
\usepackage{multirow}

\usepackage{amsfonts}
\usepackage{amssymb}
\usepackage{bbm}
\usepackage{comment}
\usepackage{textcomp}
\usepackage{array}
\usepackage{calc}
\usepackage{subfigure}
\usepackage{xcolor}
\usepackage{color}
\usepackage{enumitem}
\usepackage{hhline}
\usepackage{enumitem}
\usepackage{xspace}
\usepackage{dblfloatfix}

\usepackage{float}

\newcommand{\etc}{\textit{etc.}\xspace}
\newcommand{\ie}{\textit{i.e.,}\xspace}
\newcommand{\eg}{\textit{e.g.,}\xspace}

\newcommand{\aka}{\textit{a.k.a.}\xspace}

 \DeclareMathOperator*{\argmin}{argmin}

\ifCLASSOPTIONcompsoc
  \usepackage[nocompress]{cite}
\else
  \usepackage{cite}
\fi
\ifCLASSINFOpdf
  \usepackage[pdftex]{graphicx}
  \usepackage{multirow}
  \usepackage{amsmath}
  \usepackage{mathtools}
\else
\fi
\begin{document}
%
\title{\textsc{AugLoss}: A Robust Augmentation-based Fine Tuning Methodology}


\author{ Kyle Otstot, 
\and
 Andrew Yang,
\and
 John Kevin Cava,
\and
 Lalitha Sankar}
 
\author{
    \IEEEauthorblockN{Kyle Otstot, Andrew Yang, John Kevin Cava,  Lalitha Sankar}\\
    \vspace{0.02in}
    \IEEEauthorblockA{Arizona State University, Tempe, Arizona, 85281 \\
    \vspace{0.02in}
    \{kotstot, ahyang2, jcava, lsankar\}@asu.edu}
}



%


\maketitle

\begin{abstract}
Deep Learning (DL) models achieve great successes in many domains. However, DL models increasingly face safety and robustness concerns, including noisy labeling in the training stage and feature distribution shifts in the testing stage. Previous works made significant progress in addressing these problems, but the focus has largely been on developing solutions for only one problem at a time. For example, recent work has argued for the use of tunable robust loss functions to mitigate label noise, and data augmentation (\eg \textsc{AugMix}) to combat distribution shifts. As a step towards addressing both problems simultaneously, we introduce \textsc{AugLoss}, a simple but effective \textit{methodology} that achieves robustness against both train-time noisy labeling and test-time feature distribution shifts by unifying data augmentation and robust loss functions. We conduct comprehensive experiments in varied settings of real-world dataset corruption to showcase the gains achieved by \textsc{AugLoss} compared to previous state-of-the-art methods. Lastly, we hope this work will open new directions for designing more robust and reliable DL models under real-world corruptions.
The GitHub links to the paper and code repository are: \url{https://github.com/SankarLab/AugLoss} and \url{https://github.com/SankarLab/AugLoss-Tiny-Imagenet}.
\end{abstract}


%
\IEEEpeerreviewmaketitle

\section{Introduction}
\IEEEPARstart{D}{eep} learning (DL) models achieve great successes in many domains. 
With such great successes, DL models have been deployed in many applications, even safety-critical applications (\eg autonomous driving and healthcare). Modern DL models rely heavily on the ability of training data to estimate and represent the data encountered during deployment. However, such a design introduces problems in both the training and testing stages of the machine learning (ML) pipeline. In the training stage, data curating procedures are often imperfect, leading to errors in the  labeling process. For instance, recent results highlight that the amount of label noise in publicly-available image datasets is $\approx 8\%$ to $38.5\%$~\cite{pmlr-v97-song19b}. This, in turn, affects the robustness and reliability of the DL models. 
%

In the testing stage, the trained and deployed model often encounters new scenarios, thereby introducing a mismatch between the train and test distributions. State-of-the-art (SoTA) DL models exhibit overconfident predictions when the train and test sets are independent and identically distributed \cite{wang2021generalizing}, which leaves them particularly vulnerable to test-time feature distribution shifts \cite{quinonero2008dataset}. For instance, \cite{hendrycks2019benchmarking} shows that adding small common corruptions (\eg snow, blue, pixelation) to the test images are enough to subvert existing classifiers; specifically, the test error of a ResNet-50 model rises from 24\% on ImageNet to 76\% on ImageNet-C.

The above two problems threaten the trustworthiness and adoption of ML algorithms in safety and security-critical domains; therefore, it is critical to develop techniques that robustify ML models under both noisy labeling in the training stage and feature distribution shift in the testing stage. There has been steady progress made in addressing these problems, but focus has largely been on developing solutions for only one problem at a time. 
For instance, a range of methods including meta-learning \cite{zheng2021meta,zhang2020distilling}, loss correction \cite{patrini2017making}, and parameterized loss functions~\cite{Wang_Wen_Pan_Zhang_2021,ma2020normalized,zhang2018generalized,sypherd2022tunable} have been proposed to robustify models against noisy labeling. However, it is unclear that such methods may continue to be robust in the face of a test-time distribution shift. 
Similarly, many methods have been shown to improve a model's generalization ability against feature distribution shift, including data augmentation \cite{hendrycks2020many,madry2017towards}, adversarial learning \cite{wang2021augmax}, and pre-training \cite{jiang2020robust,chen2020adversarial}. 
However, such methods have only been considered when the dataset is exclusively corrupted by distribution shifts; their ability to handle noisy training labels are unknown and unexplored. 

In this work, we highlight the necessity of considering \textit{both} problems simultaneously while designing robust DL models for real-world systems. In doing so, we present a robust methodology to tackle corruptions in both training (\eg noisy labeling) and testing (\eg feature distribution shift) stages. To this end, we select two representative methods, namely, robust (parameterized) loss functions and data augmentation, that have been often used to address training- and testing-stage corruptions, respectively. Unifying data augmentation techniques with robust loss functions, we propose \textsc{AugLoss} as a simple, yet effective \textit{methodology} to enhance robustness against both train-time noisy labeling and test-time distribution shifts. 

We consider three realizations of \textsc{AugLoss} by pairing each of three well-studied robust loss functions --  focal loss \cite{focal2017}, normalized cross entropy + reverse cross entropy~(NCE+RCE) loss \cite{ma2020normalized}, and $\alpha$-loss \cite{sypherd2022tunable} -- with \textsc{AugMix} data augmentation \cite{hendrycks2019augmix}. For comparison with oft-used methods, our experiments also include the following: (i) cross entropy (CE) loss without \textsc{AugMix}, (ii) \textsc{AugMix} without a robust loss (\ie with CE), (iii) each of the three robust losses without \textsc{AugMix}. We evaluate all of these options for augmentation-loss choices by conducting comprehensive experiments with the CIFAR-10, CIFAR-100, and Tiny ImageNet datasets. For each realization, we produce WideResNet \cite{zagoruyko2016wide} models for CIFAR-10 and CIFAR-100 datasets and EfficientNet \cite{mingxing2021efficient} models for the Tiny ImageNet dataset by first training on varied settings of synthetic and human-annotated label noise \cite{wei2021learning}, then testing on corrupted images of CIFAR-10, CIFAR-100, and Tiny ImageNet \cite{hendrycks2019benchmarking}. 
We summarize our key contributions below. 



\begin{itemize}[leftmargin=*]
    \item To the best of our knowledge, 
    our proposed \textit{methodology}, \textsc{AugLoss}, is the first to combine data augmentation and robust loss functions, thus offering an effective solution to train-time noisy labeling and test-time distribution shift. More broadly, \textsc{AugLoss} provides a blueprint to study the efficacy of different augmentation and loss combinations.
    \item 
    Our comprehensive experiments in varied settings of train-label and test-feature corruptions showcase the gains achieved by \textsc{AugLoss} compared to all other augmentation-loss combinations listed earlier.
    Our key contribution is in highlighting the efficacy of different combinations for different corruption scenarios. In particular, we show that \textsc{AugLoss} (\textsc{AugMix} + robust loss) methods consistently outperform the other combinations across every tested setting of label noise.  
    \item Lastly, we observe that although \textsc{AugLoss} outperforms SoTA methods, our results also suggest that no single robust loss function within the \textsc{AugLoss} framework is a universal "best fit" across all tested settings of label noise. In particular, our results suggest that the focal loss performs well on CIFAR-100 corrupted by asymmetric noise, the NCE+RCE loss generally performs well under settings of high label noise (0.3,0.4), and the $\alpha$-loss performs well on datasets corrupted by symmetric label noise. 
    %
    
\end{itemize}

Our results, detailed in the sequel, highlight that our proposed \textsc{AugLoss} methodology can enhance the performance of DL models under both train-time noisy labeling and test-time feature distribution shifts. We believe this is an important step toward real-world robust DL systems. The rest of the paper is organized as follows. In Section \ref{section:relatedwork}, we review the literature of robust loss functions and data augmentation techniques. In Section \ref{section:preliminaries}, we present the background, problem setup, and formulation of both robust loss and data augmentation. Section \ref{section:augloss} outlines the \textsc{AugLoss} framework, and Section \ref{section:experiments} details our experimental setup, results, and discussion. Finally, we conclude in Section \ref{section:conclusion}.

\section{Related Work}

\label{section:relatedwork}

\subsection{Loss Functions for Noisy Labeling}
Using loss functions to mitigate noisy labels during training is a strong tradition in machine learning~\cite{natarajan2013learning,ghosh2017robust,patrini2017making}. 
A common approach is the enhancement of the cross entropy loss with hyperparameters (and possibly additional regularizing terms) that enable the practitioner to tailor the loss to the desired application. 
In this vein, an important hyperparameterized loss is the generalized cross entropy (GCE), which was motivated by the Box-Cox transformation in statistics~\cite{zhang2018generalized}.
GCE was experimentally shown to be robust to noisy labels in neural-networks.
A generalization of GCE, called $\alpha$-loss, was motivated by Arimoto entropies arising in information theory~\cite{sypherd2022tunable}.
The $\alpha$-loss was experimentally shown to be robust to noisy labels during training across several algorithms (neural-networks, logistic regression, and boosting), along with theoretical support of these robustness characteristics~\cite{properlyimproper2022}.
Another important hyperparameterized loss is the focal loss~\cite{focal2017}.
While the focal loss was initially used for dense object detection with great success, it has recently received deeper theoretical scrutiny~\cite{charoenphakdee2021focal,properlyimproper2022}, 
been observed to improve neural-network calibration~\cite{mukhoti2020calibrating}, and also applied to the problem of noisy labels~\cite{ma2020normalized}.
Lastly, an important hyperparameterized loss function for noisy labels is NCE+RCE, which stems from the framework of active-passive losses~\cite{ma2020normalized}.
Unlike $\alpha$-loss and focal loss, NCE+RCE employs two hyperparameters and is a linear combination of losses (see Section~\ref{subsec:robustlossfunctions} for more details). 
Important for our purposes, NCE+RCE has been successfully employed in computer vision applications, particularly in the very high noise regime. 
Overall, $\alpha$-loss, focal loss, and NCE+RCE have all been shown to be robust to label noise in the training data, and hence comprise a strong representative subset of the robust loss function literature (for more examples, see~\cite{wang2019symmetric,zhou2021asymmetric,liu2020peer,leng2022polyloss}).
However, to the best of our knowledge, each of these loss functions have not been previously considered in the joint setting of training \textit{and} test domain shift, which we argue is the real-world scenario addressed by our proposed \textsc{AugLoss} methodology.


\subsection{Data Augmentation for Domain Adaptation}

The issue of domain shifts is an active field. In particular, DL models learning on train data may not generalize well to the test data if the train and test distributions are misaligned. For example, adding small corruptions (\eg snow, blue, pixelation) to test images nearly triples the generalization error of ResNet-50 models training on ImageNet \cite{hendrycks2019benchmarking}. Consequently, the field of domain adaption has received much attention, including the proposal of data augmentation techniques that help robustify models under domain shift. Specifically, Cutout, the occlusion of square input regions during training, was shown to improve generalization error on clean images \cite{devries2017cutout}. Building on this, CutMix replaces the occluded regions with portions of different images, and yields better test performance on out-of-distribution examples \cite{yun2019cutmix}. Similarly, Mixup synthesizes information from two images by training on convex combinations of two feature-target pairs \cite{zhang2017mixup}. Taking a different approach, AutoAugment searches for improved data augmentation policies -- successive operations (\eg translation, rotation) that preserve image semantics -- achieving state-of-the-art validation accuracy on clean datasets \cite{cubuk2018autoaugment}. Combining the previous two approaches, \textsc{AugMix} enhances the training set with mixtures of three (or less) augmented images, each generated by its own chain of stochastically-sampled operations \cite{hendrycks2019augmix}. Out of the proposed data augmentation methods, \textsc{AugMix} achieves the best performance for out-of-distribution testing; however, descendants of \textsc{AugMix}, namely AugMax \cite{wang2021augmax} and NoisyMix \cite{erichson2022noisymix}, combine data augmentation with adversarial learning and stability training, respectively, to achieve even better results under domain shift. In this work, we consider the unification of data augmentation with robust loss functions -- one that, to our best knowledge, is unexplored.

\section{Preliminaries}

\label{section:preliminaries}

\subsection{Image Classification}

Consider a feature-label pair of random variables $(X,Y) \sim q_{X,Y}$ sampled over feature space $\mathcal{X}$ and label space $\mathcal{Y}$. In this, we acknowledge the existence of a ground-truth posterior distribution $q_{Y | X}$ where $q_{Y|X}(y|x)$ is the probability that feature $x \in \mathcal{X}$ is representative of label $y \in \mathcal{Y}$, and a ground-truth marginal (or prior) distribution $q_{X}$ over the the feature space, where $q_{X}(x)$ is the probability that we observe $x \in \mathcal{X}$. 
The underlying distribution $q_{X,Y}$ is unknown, but we have access to some dataset serving as a finite representation of $q_{X,Y}$. We then look to develop a method that learns a best estimate $\hat{q}_{Y | X}$ of the posterior $q_{Y | X}$ given the observed dataset. An effective method in this context should work to minimize the difference between $\hat{q}_{Y | X}$ and $q_{Y | X}$.

Oftentimes, the inner workings of the data-generating process are inaccessible, and consequently we are only provided the outcome (\ie the observed dataset), which has encouraged the development of practical methods biased to several convenient assumptions about the dataset. For example, it is common to assume that all dataset samples are i.i.d: independent, and \textit{identically distributed} according to the evaluation samples. This way, a model fitting on a subset of the data can generalize to the unforeseen examples reflective of the ground truth. However, the data-generating process is susceptible to particular obstacles that prevent the satisfaction of i.i.d. and consequently hinder the efficacy of learning methods depending on these assumptions. Unaddressed limitations of data collection can mislead the model's perspective of $q_{Y|X}$ and $q_{X}$ -- the two factors composing $q_{X,Y}$ -- and ultimately welcome data drawn from some perturbation of the ground-truth. We note these limitations below.

\subsubsection{Mislabeling of the Sampled Features}

Supervised learning requires each observed feature in the dataset to be properly annotated, but the accuracy is often mitigated by factors including the subversion of experienced domain experts \cite{frenay2013classification}, the unreliable nature of crowdsourcing platforms \cite{arpit2017closer}, and the threat of adversarial label-flip attacks \cite{xiao2012adversarial}. As a result, large-scale datasets are prone to noisy labeling, where a fraction of the true labels are flipped to false classes. In this case, the train labels are drawn from some perturbation $\tilde{q}_{Y|X}$ of the true posterior $q_{Y|X}$.

In response to the growing concern of noisy labeling, recent work has been done on synthetic label noise generation within clean datasets \cite{zhang2017understanding}. Standard methods consider both symmetric (random labels flipped to other uniform-random classes) and asymmetric (random labels flipped to other visually-similar classes) approaches to noise generation. Since label noise is frequently associated with human error, real-world examples of asymmetric generation have recently received much interest: CIFAR-10N and CIFAR-100N (\aka CIFAR-*N), for instance, contain human-annotated noisy labeling for the CIFAR-10 and CIFAR-100 images, collected by Amazon Mechanical Turk \cite{wei2021learning}. 

DL models that exhibit robustness under label corruption (noise) should give accurate classifications even after fitting on a dataset with noisy labeling. Therefore, the standard approach to evaluating a model’s robustness includes: (1) training on noisy labels, generated symmetrically or asymmetrically; and (2) testing on clean labels.

\subsubsection{Imperfect Sampling of the Feature Space}

The data collection process requires each image in the dataset to be sampled randomly from the feature space. However, the execution of random sampling in practice is often challenged by several factors. For example, the feature-generating distribution may evolve over time and experience a distributional shift during deployment \cite{wang2021generalizing}. Additionally, image datasets tend to reflect built-in biases in which the curated features are drawn from a small subset of the underlying distribution \cite{patrini2017making}. In either case, a distribution mismatch exists between the training and testing stages, and as a result, the train features are sampled from a perturbation $\tilde{q}_{X}$ of the true prior $q_{X}$.

\begin{figure}
\centering
\includegraphics[width=8cm]{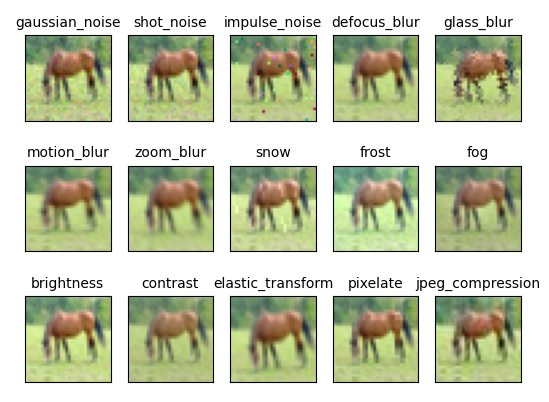}
\caption{The 15 common corruptions on a HORSE-labeled image, found in the CIFAR-10-C dataset.}
\label{fig:cfc}
\end{figure}

Recent work has been done to address scenarios of unforeseen data shifts in the prior distribution, including the generation of real-world feature noise on clean datasets. CIFAR-10-C and CIFAR-100-C (\aka CIFAR-*C), as well as Tiny ImageNet-C for instance, use fifteen algorithmically-generated corruptions -- including real-world noise, blur, weather, and digital categories -- on the CIFAR-10, CIFAR-100, and Tiny ImageNet datasets to generate a new image set reflective of data shifts potentially encountered in practice \cite{hendrycks2019benchmarking}. Figure \ref{fig:cfc} shows the fifteen corruptions on a CIFAR-10 image.

The classification ability of robust, real-world DL models should generalize to corrupted features after fitting on a clean dataset. Hence, the standard approach to evaluating a model’s robustness under data shift should include: (1) training on clean features; and (2) testing on noisy features, generated by common corruptions.

\subsection{Dataset Corruption \& Proposed Remedies}

In this paper, we consider a dataset to be corrupted if its train labels and features are drawn from a misaligned posterior $\tilde{q}_{Y|X}$ and misaligned prior $\tilde{q}_{X}$, respectively. With the increasing prevalence of such large-scale datasets, we argue that errors in the data collection process are inevitable, so effective learning methods require a particular degree of robustness under realistic settings of dataset corruption. Our methodology is primarily motivated by previous work done on model robustness under label noise or feature noise, but not both. Specifically, we consider two proposed techniques that robustify models under one of the two noise types. The techniques are presented below.

\begin{figure*}
\centering
\includegraphics[width=18cm]{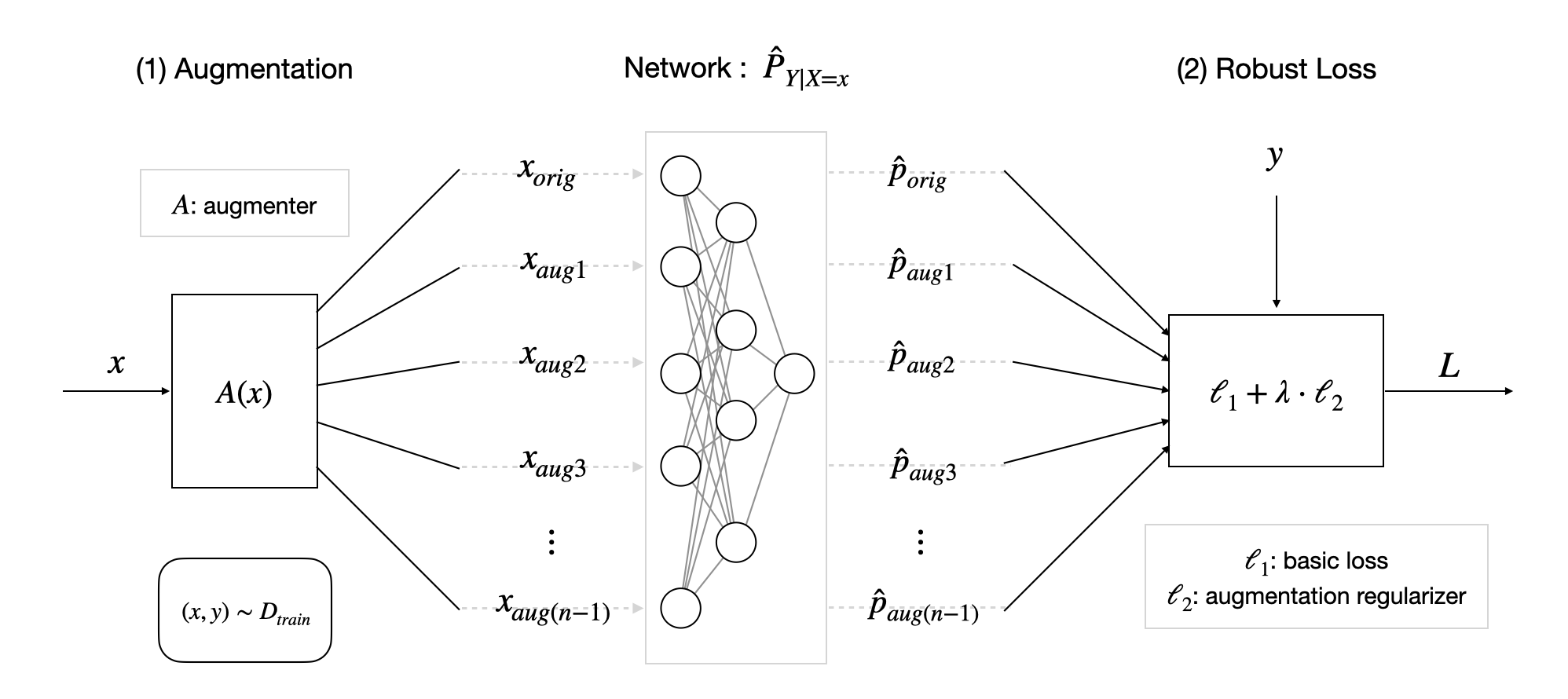}
\caption{\textsc{AugLoss}: the unification of data augmentation and robust loss functions.}
\label{fig:augloss}
\end{figure*}

\subsubsection{Robust Loss Functions} \label{subsec:robustlossfunctions}

The basic loss function $\ell : \mathcal{P}(Y|X) \times \mathcal{Y} \rightarrow \mathbb{R}^{+}$ maps an estimated posterior distribution and true class label to a performance measure with the intent of learning a posterior distribution $\hat{p}$ that minimizes the expectation of $\ell\left( \hat{p}(x), y\right)$ across all $(x,y)$ pairs in the train set. For any reasonable label corruption $D_{\text{train}} \mapsto \tilde{D}_{\text{train}}$ to the train set, we generally expect a loss function $\ell$ to be \textit{robust} when the following property holds:
\begin{equation}
\argmin_{\hat{p}}\mathbb{E}_{\tilde{D}_{\text{train}}} \ell\left( \hat{p}(x), y\right) \approx \argmin_{\hat{p}}\mathbb{E}_{D_{\text{train}}} \ell\left( \hat{p}(x), y\right).
\end{equation}
However, in this paper, we relax the intuitive representation of "robust loss" by asserting the following: loss functions exhibit robustness if they achieve better performance, in settings of train-time noisy labeling, than the standard loss function in image classification, cross entropy (CE)
\begin{equation}
\ell_{\text{CE}}(\hat{p}, y) = - \log \left( \hat{p}(y)\right),
\end{equation}
which has been shown to be non-robust under label noise~\cite{ma2020normalized}. Moreover, recent progress has been made in the formulation of robust \textit{tunable loss function families}, including focal loss \cite{lin2018focal}, active-passive loss \cite{ma2020normalized}, and $\alpha$-loss \cite{sypherd2022tunable}. Specifically, a loss function family $\ell_{\theta}$ parameterized by $\theta$ in the space $\Theta$ is considered robust when there exists a tuning $\theta^{*} \in \Theta$ such that $\ell_{\theta^{*}}$ is robust.

The \textbf{focal loss} family parameterized by $\gamma \in [0,5]$,
\begin{equation}
    \ell_{\text{FL}}(\hat{p}, y; \gamma) = -\left(1 - \hat{p}(y)\right)^{\gamma} \log \left( \hat{p}(y)\right),
\end{equation}
is experimentally shown to perform better than CE loss under label noise when $\gamma$ is tuned above $0$ \cite{lin2018focal} ($\gamma = 0$ is equivalent to CE). The \textbf{NCE+RCE loss} family parameterized by $(\beta_{1},\beta_{2}) \in \mathbb{R}_{+}^{2}$,
\begin{equation}
    \ell_{\text{NCE+RCE}}(\hat{p}, y; \beta_{1}, \beta_{2}) = \beta_{1} \cdot \ell_{\text{NCE}}(\hat{p}, y) + \beta_{2} \cdot \ell_{\text{RCE}}(\hat{p}, y),
\end{equation}
is an example of active-passive loss -- a proposed group of loss function families that linearly combine an \textit{active} loss function and \textit{passive} loss function. In this case, the active loss is normalized cross entropy (NCE)
\begin{equation}
   \ell_{\text{NCE}}(\hat{p}, y) = \frac{\ell_{\text{CE}}(\hat{p}, y)}{\sum_{y' \in \mathcal{Y}} \ell_{\text{CE}}(\hat{p}, y')},
\end{equation}
and the passive loss is reverse cross entropy (RCE)
\begin{equation}
\ell_{\text{RCE}}(\hat{p}, y; \delta) = \delta \sum_{y' \neq y} \hat{p}(y') \text{  }, \text{   } \delta > 0 .
\end{equation}
In the NCE+RCE paper, $\delta$ is fixed to $4$. Lastly, the $\boldsymbol\alpha$\textbf{-loss} family parameterized by $\alpha \in (0, \infty]$, \begin{equation}
\ell_{\alpha}(\hat{p}, y; \alpha) = \frac{\alpha}{\alpha - 1} \left( 1 - \hat{p}(y)^{1 - 1/\alpha} \right),
\end{equation}
encapsulates the exponential ($\alpha = 1/2$), cross entropy ($\alpha = 1$), and 0-1 ($\alpha \rightarrow \infty$) losses. This family is shown to perform better than CE loss under label noise when $\alpha > 1$ \cite{sypherd2022tunable}.

\subsubsection{Data Augmentation}

Motivated by previous work on data augmentation~\cite{hendrycks2020augmix,huang2021dair}, we generalize the augmenter $A : \mathcal{X} \rightarrow \mathcal{X}^{n}$ to return an $n$-tuple 
\begin{equation}
A(x) = (x_{\text{orig}}, x_{\text{aug}1}, x_{\text{aug}2},\dots, x_{\text{aug}(n-1)})
\end{equation}
from a given feature $x \in \mathcal{X}$, where $x_{\text{orig}} := x$ and each $x_{\text{aug}(i)}$ is a unique transformation of $x$. The model then learns $n$ distinct distributions 
\begin{equation}
\hat{P} := \left(\hat{p}_{\text{orig}}, \hat{p}_{\text{aug}1}, \hat{p}_{\text{aug}2}, \dots, \hat{p}_{\text{aug}(n-1)}\right),
\end{equation}
where $\hat{p}_{\text{orig}}$ fits on the train set $D_{\text{train}}$, and each $\hat{p}_{\text{aug}(i)}$ fits on the dataset $\{(x_{\text{aug}(i)}, y) : (x, y) \in D_{\text{train}}\}$.

The use of augmentation warrants a loss supplement $\ell_{2} : \mathcal{P}(Y|X)^{n} \rightarrow \mathbb{R}^{+}$ that synthesizes the information of each learned distribution in $\hat{P}$ into a measure of similarity. Assuming true class preservation within $A(x)$, an effective form of regularization will ensure that $\forall_{i \in [n-1]} \text{ } \hat{p}_{\text{aug}(i)} \approx \hat{p}_{\text{orig}}$, which aims to improve the model's robustness under varied settings of feature corruption. Training with data augmentation makes use of the general loss function $\mathcal{L}$
\begin{equation} \label{eq:loss}
\mathcal{L}(\hat{P}, y; \lambda) = \ell_{1}(\hat{p}_{\text{orig}}, y) + \lambda \cdot \ell_{2}(\hat{P}),
\end{equation}
where $\ell_{1}$ is a basic loss function (\eg CE loss) and $\ell_{2}$ is an augmentation regularizer scaled by some constant $\lambda \in \mathbb{R}^{+}$.

\section{\textsc{AugLoss} Framework}

\label{section:augloss}

Previously, we made note of the following:
\begin{enumerate}
    \item Robustness under \textit{label} corruption should be evaluated by training on noisy labels, then testing on clean labels.
    \item Robustness under \textit{feature} corruption should be evaluated by training models on clean features, then testing them on noisy features.
\end{enumerate}
Motivated by robustness under real-world dataset corruption, we choose to unify the above two statements and propose that robust DL models should perform well on the \textit{novel task} of training on noisy labels and clean features, followed by testing on clean labels and noisy features. In this section, we introduce \textsc{AugLoss}: a learning methodology that combines the known techniques of data augmentation and robust loss functions, formulating an effective solution to our novel task. The two stages of this methodology are illustrated in Figure \ref{fig:augloss} and outlined below.

\subsection*{Stage 1: Classification}
The classification stage begins with a random sampling $(x,y) \in D_{\text{train}}$ of the train set; in practice, samples are done in batches, but for clarity of exposition we notate a single example. In the first step, data augmentation is performed with an augmenter-of-choice $A$. The augmenter runs the original image $x$ through a number of distinct transformations, returning $x_{\text{orig}} := x$ along with the newly-transformed images. Appropriate transformations for the task-at-hand should be nontrivial: $x_{\text{aug}(i)} \neq x_{\text{orig}}$; uniquely generated: $x_{\text{aug}(i)} \neq x_{\text{aug}(j)}, \forall_{i \neq j}$; visually realistic:  $q_{X = x_{\text{aug}(i)}} \approx q_{X=x_{\text{orig}}}$; preservative of class representations: $q_{Y | X = x_{\text{aug}(i)}} \approx q_{Y | X=x_{\text{orig}}}$; and computationally feasible.

Then, each image in $A(x)$ is fed through a network of sufficient capacity (relative to the dataset) to produce a tuple of estimated posterior distributions. Each output $\hat{p}_{\text{aug}(i)}$ represents a distribution over the label space, conditioned on the augmented feature $x_{\text{aug}(i)}$.


\subsection*{Stage 2: Evaluation \& Correction}

The transformations undergone in the augmentation stage should work to preserve the true class representation of the original image, so an accurate classifier should embed each augmented image similarly to the original one. To ensure this, we train the model with the loss function $\mathcal{L}$ in Eq. \ref{eq:loss}, pairing a basic loss function $\ell_{1}(\hat{p}_{\text{orig}}, y)$ with an augmentation regularizer $\ell_{2}(\hat{P})$ that is minimized when the distributions in $\hat{P}$ are all the same. We set $\ell_{1}$ to be an optimally-tuned instance of a robust loss function family.

As a result, the loss function takes the form $\mathcal{L} = \ell_{1} + \lambda \cdot \ell_{2}$, where (1) the loss $\ell_{1}$ works to robustify the model against label noise in the training stage, and (2) the regularizer $\ell_{2}$ works to generalize the performance of the model to unforeseen data shifts in the testing stage. The regularization constant $\lambda$ is fixed to a value that helps $\mathcal{L}$ balance between these two objectives. The optimization stage of the model searches for an estimated posterior distribution
\begin{equation}\hat{p}^{*} := \argmin_{\hat{p}} \mathbb{E}_{D_\text{train}} \mathcal{L}\left((\hat{p}_{\text{orig}}, \hat{p}_{\text{aug}1}, ...), y; \lambda\right)\end{equation}
by learning (on a noisy-labeled train set) the class representations invariant to common image corruptions.


\begin{table*}[h!]
\refstepcounter{table}
\label{table:mce-sym10}
\centering
\renewcommand{\arraystretch}{0.9}
\resizebox{\textwidth}{!}{\begin{tabular}{@{}ccclclclclclcl@{}}
\toprule
\multicolumn{2}{c}{Method} &
  \multicolumn{10}{c}{TABLE \ref{table:mce-sym10}: \textbf{Symmetric} CIFAR-10} \\ \midrule
Augment &
  Loss &
  \multicolumn{2}{c}{$\eta$ = 0} &
  \multicolumn{2}{c}{0.1} &
  \multicolumn{2}{c}{0.2} &
  \multicolumn{2}{c}{0.3} &
  \multicolumn{2}{c}{0.4} &
  \multicolumn{2}{c}{Noisy Avg.} \\ \midrule
\multirow{4}{*}{NoAug} &
  CE &
  \multicolumn{2}{c}{$\mathbf{27.57 \pm 0.61}$} &
  \multicolumn{2}{c}{$35.79 \pm 0.62$} &
  \multicolumn{2}{c}{$40.45 \pm 1.37$} &
  \multicolumn{2}{c}{$45.19 \pm 1.06$} &
  \multicolumn{2}{c}{$50.94 \pm 0.34$} &
  \multicolumn{2}{c}{$43.09 \pm 0.85$} \\ \cmidrule(l){2-14} 
 &
  Focal &
  \multicolumn{2}{c}{$33.81 \pm 2.05$} &
  \multicolumn{2}{c}{$32.58 \pm 0.95$} &
  \multicolumn{2}{c}{$37.98 \pm 1.06$} &
  \multicolumn{2}{c}{$44.15 \pm 0.22$} &
  \multicolumn{2}{c}{$50.47 \pm 0.19$} &
  \multicolumn{2}{c}{$41.30 \pm 0.61$} \\ \cmidrule(l){2-14} 
 &
 
  NCE+RCE &
  \multicolumn{2}{c}{$30.02 \pm 0.30$} &
  \multicolumn{2}{c}{$30.42 \pm 0.13$} &
  \multicolumn{2}{c}{$31.80 \pm 1.20$} &
  \multicolumn{2}{c}{$34.20 \pm 0.24$} &
  \multicolumn{2}{c}{$35.96 \pm 1.43$} &
  \multicolumn{2}{c}{$33.10 \pm 0.75$} \\ \cmidrule(l){2-14}
 &
 
  $\alpha$-loss &
  \multicolumn{2}{c}{$28.43 \pm 0.47$} &
  \multicolumn{2}{c}{$\mathbf{29.28 \pm 1.21}$} &
  \multicolumn{2}{c}{$\mathbf{30.23 \pm 0.81}$} &
  \multicolumn{2}{c}{$\mathbf{32.22 \pm 0.75}$} &
  \multicolumn{2}{c}{$\mathbf{35.16 \pm 0.51}$} &
  \multicolumn{2}{c}{$\mathbf{31.72 \pm 0.82}$} \\ \midrule
  
\multirow{4}{*}{\textsc{AugMix}} &
  CE &
  \multicolumn{2}{c}{$12.62 \pm 0.40$} &
  \multicolumn{2}{c}{$15.14 \pm 0.25$} &
  \multicolumn{2}{c}{$18.02 \pm 0.11$} &
  \multicolumn{2}{c}{$21.23 \pm 0.23$} &
  \multicolumn{2}{c}{$26.46 \pm 0.59$} &
  \multicolumn{2}{c}{$20.21 \pm 0.30$} \\ \cmidrule(l){2-14} 
 &
  Focal &
  \multicolumn{2}{c}{$13.81 \pm 0.21$} &
\multicolumn{2}{c}{$\mathbf{12.59 \pm 0.05}$} &
\multicolumn{2}{c}{$13.91 \pm 0.2$} &
\multicolumn{2}{c}{$17.79 \pm 0.46$} &
\multicolumn{2}{c}{$23.15 \pm 0.33$} &
\multicolumn{2}{c}{$16.86 \pm 0.26$} \\ \cmidrule(l){2-14}
 &

  NCE+RCE &
  \multicolumn{2}{c}{$13.15 \pm 0.16$} &
  \multicolumn{2}{c}{$13.72 \pm 0.40$} &
  \multicolumn{2}{c}{$13.87 \pm 0.09$} &
  \multicolumn{2}{c}{$14.19 \pm 0.23$} &
  \multicolumn{2}{c}{$15.17 \pm 0.12$} &
  \multicolumn{2}{c}{$14.24 \pm 0.21$} \\ \cmidrule(l){2-14}
  
 &

  $\alpha$-loss &
  \multicolumn{2}{c}{$\mathbf{12.55 \pm 0.09}$} &
  \multicolumn{2}{c}{$12.82 \pm 0.09$} &
  \multicolumn{2}{c}{$\mathbf{13.23 \pm 0.13}$} &
  \multicolumn{2}{c}{$\mathbf{13.81 \pm 0.07}$} &
  \multicolumn{2}{c}{$\mathbf{15.10 \pm 0.27}$} &
  \multicolumn{2}{c}{$\mathbf{13.74 \pm 0.14}$} \\ \bottomrule
  
\end{tabular}}

\end{table*}

\begin{table*}[h!]
\refstepcounter{table}
\label{table:mce-sym100}
\centering
\renewcommand{\arraystretch}{0.9}
\resizebox{\textwidth}{!}{\begin{tabular}{@{}ccclclclclclcl@{}}
\toprule
\multicolumn{2}{c}{Method} &
  \multicolumn{10}{c}{TABLE \ref{table:mce-sym100}: \textbf{Symmetric} CIFAR-100} \\ \midrule
Augment &
  Loss &
  \multicolumn{2}{c}{$\eta$ = 0} &
  \multicolumn{2}{c}{0.1} &
  \multicolumn{2}{c}{0.2} &
  \multicolumn{2}{c}{0.3} &
  \multicolumn{2}{c}{0.4} &
  \multicolumn{2}{c}{Noisy Avg.} \\ \midrule
\multirow{4}{*}{NoAug} &
  CE &
  \multicolumn{2}{c}{$53.84 \pm 0.18$} &
  \multicolumn{2}{c}{$60.22 \pm 0.35$} &
  \multicolumn{2}{c}{$64.93 \pm 0.63$} &
  \multicolumn{2}{c}{$68.68 \pm 0.23$} &
  \multicolumn{2}{c}{$72.74 \pm 0.51$} &
  \multicolumn{2}{c}{$66.64 \pm 0.43$} \\ \cmidrule(l){2-14} 
 &
  Focal &
  \multicolumn{2}{c}{$\mathbf{53.50 \pm 0.35}$} &
  \multicolumn{2}{c}{$57.85 \pm 0.21$} &
  \multicolumn{2}{c}{$62.73 \pm 0.46$} &
  \multicolumn{2}{c}{$67.18 \pm 0.35$} &
  \multicolumn{2}{c}{$72.35 \pm 0.49$} &
  \multicolumn{2}{c}{$65.03 \pm 0.38$} \\ \cmidrule(l){2-14} 
 &
 
  NCE+RCE &
  \multicolumn{2}{c}{$54.66 \pm 0.38$} &
  \multicolumn{2}{c}{$\mathbf{55.34 \pm 0.54}$} &
  \multicolumn{2}{c}{$56.91 \pm 0.16$} &
  \multicolumn{2}{c}{$58.76 \pm 0.60$} &
  \multicolumn{2}{c}{$61.92 \pm 0.67$} &
  \multicolumn{2}{c}{$58.23 \pm 0.49$} \\ \cmidrule(l){2-14}
 &
 
  $\alpha$-loss &
  \multicolumn{2}{c}{$54.43 \pm 0.13$} &
  \multicolumn{2}{c}{$55.58 \pm 0.24$} &
  \multicolumn{2}{c}{$\mathbf{56.48 \pm 0.40}$} &
  \multicolumn{2}{c}{$\mathbf{58.11 \pm 0.61}$} &
  \multicolumn{2}{c}{$\mathbf{60.21 \pm 0.81}$} &
  \multicolumn{2}{c}{$\mathbf{57.60 \pm 0.52}$} \\ \midrule
  
\multirow{4}{*}{\textsc{AugMix}} &
  CE &
  \multicolumn{2}{c}{$37.80 \pm 0.22$} &
  \multicolumn{2}{c}{$42.11 \pm 0.19$} &
  \multicolumn{2}{c}{$44.86 \pm 0.50$} &
  \multicolumn{2}{c}{$48.34 \pm 0.18$} &
  \multicolumn{2}{c}{$51.65 \pm 0.09$} &
  \multicolumn{2}{c}{$46.74 \pm 0.24$} \\ \cmidrule(l){2-14} 
 &
  Focal &
  \multicolumn{2}{c}{$\mathbf{36.11 \pm 0.2}$} &
\multicolumn{2}{c}{$38.55 \pm 0.47$} &
\multicolumn{2}{c}{$42.5 \pm 0.26$} &
\multicolumn{2}{c}{$46.97 \pm 0.05$} &
\multicolumn{2}{c}{$52.05 \pm 0.09$} &
\multicolumn{2}{c}{$45.02 \pm 0.22$} \\ \cmidrule(l){2-14}
 &

  NCE+RCE &
  \multicolumn{2}{c}{$41.53 \pm 0.96$} &
  \multicolumn{2}{c}{$42.18 \pm 0.10$} &
  \multicolumn{2}{c}{$42.84 \pm 0.14$} &
  \multicolumn{2}{c}{$42.71 \pm 0.46$} &
  \multicolumn{2}{c}{$\mathbf{44.17 \pm 0.80}$} &
  \multicolumn{2}{c}{$42.98 \pm 0.38$} \\ \cmidrule(l){2-14}
  
 &

  $\alpha$-loss &
  \multicolumn{2}{c}{$37.66 \pm 0.13$} &
  \multicolumn{2}{c}{$\mathbf{38.46 \pm 0.26}$} &
  \multicolumn{2}{c}{$\mathbf{40.03 \pm 0.50}$} &
  \multicolumn{2}{c}{$\mathbf{41.90 \pm 0.51}$} &
  \multicolumn{2}{c}{$44.54 \pm 0.30$} &
  \multicolumn{2}{c}{$\mathbf{41.23 \pm 0.39}$} \\ \bottomrule
  
\end{tabular}}

\end{table*}

\begin{table*}[h!]
\refstepcounter{table}
\label{table:mce-timgnet}
\centering
\renewcommand{\arraystretch}{0.9}
\resizebox{\textwidth}{!}{\begin{tabular}{@{}ccclclclclclcl@{}}
\toprule
\multicolumn{2}{c}{Method} &
  \multicolumn{10}{c}{TABLE \ref{table:mce-timgnet}:
  \textbf {Symmetric} Tiny ImageNet}\\ \midrule
Augment &
  Loss &
  \multicolumn{2}{c}{$\eta$ = 0} &
  \multicolumn{2}{c}{0.1} &
  \multicolumn{2}{c}{0.2} &
  \multicolumn{2}{c}{0.3} &
  \multicolumn{2}{c}{0.4} &
  \multicolumn{2}{c}{Noisy Avg.} \\ \midrule
\multirow{4}{*}{NoAug} &
  CE &
  \multicolumn{2}{c}{$50.72 \pm 0.32$} &
  \multicolumn{2}{c}{$52.31 \pm 0.18$} &
  \multicolumn{2}{c}{$55.23 \pm 0.26$} &
  \multicolumn{2}{c}{$60.57 \pm 0.19$} &
  \multicolumn{2}{c}{$65.29 \pm 0.16$} &
  \multicolumn{2}{c}{$56.82 \pm 0.22$} \\ \cmidrule(l){2-14} 
 &
  Focal &
  \multicolumn{2}{c}{$\mathbf{49.47 \pm 0.46}$} &
  \multicolumn{2}{c}{$\mathbf{51.18 \pm 0.21}$} &
  \multicolumn{2}{c}{$\mathbf{53.72 \pm 0.94}$} &
  \multicolumn{2}{c}{$59.12 \pm 0.44$} &
  \multicolumn{2}{c}{$64.56 \pm 0.32$} &
  \multicolumn{2}{c}{$55.61 \pm 0.47$} \\ \cmidrule(l){2-14} 
 &
  NCE+RCE &
  \multicolumn{2}{c}{$51.51 \pm 0.20$} &
  \multicolumn{2}{c}{$54.01 \pm 0.19$} &
  \multicolumn{2}{c}{$56.30 \pm 0.04$} &
  \multicolumn{2}{c}{$59.23 \pm 0.31$} &
  \multicolumn{2}{c}{$\mathbf{61.47 \pm 0.19}$} &
  \multicolumn{2}{c}{$56.50 \pm 0.19$} \\ \cmidrule(l){2-14}
 &
  $\alpha$-loss &
  \multicolumn{2}{c}{$50.64 \pm 0.41$} &
  \multicolumn{2}{c}{$52.20 \pm 0.26$} &
  \multicolumn{2}{c}{$54.22 \pm 0.11$} &
  \multicolumn{2}{c}{$\mathbf{58.08 \pm 0.14}$} &
  \multicolumn{2}{c}{$61.57 \pm 0.06$} &
  \multicolumn{2}{c}{$\mathbf{55.34 \pm 0.20}$} \\ \midrule
\multirow{4}{*}{AugMix} &
  CE &
  \multicolumn{2}{c}{$41.28 \pm 0.18$} &
  \multicolumn{2}{c}{$42.75 \pm 0.13$} &
  \multicolumn{2}{c}{$44.65 \pm 0.08$} &
  \multicolumn{2}{c}{$46.49 \pm 0.17$} &
  \multicolumn{2}{c}{$49.20 \pm 0.38$} &
  \multicolumn{2}{c}{$44.87 \pm 0.19$} \\ \cmidrule(l){2-14} 
 &
  Focal &
  \multicolumn{2}{c}{$\mathbf{39.90 \pm 0.24}$} &
  \multicolumn{2}{c}{$41.66 \pm 0.26$} &
  \multicolumn{2}{c}{$43.43 \pm 0.14$} &
  \multicolumn{2}{c}{$44.99 \pm 0.24$} &
  \multicolumn{2}{c}{$48.14 \pm 0.24$} &
  \multicolumn{2}{c}{$43.62 \pm 0.22$} \\ \cmidrule(l){2-14}
 &
  NCE+RCE &
  \multicolumn{2}{c}{$41.73 \pm 0.53$} &
  \multicolumn{2}{c}{$42.84 \pm 0.09$} &
  \multicolumn{2}{c}{$43.88 \pm 0.26$} &
  \multicolumn{2}{c}{$44.78 \pm 0.40$} &
  \multicolumn{2}{c}{$46.39 \pm 0.16$} &
  \multicolumn{2}{c}{$43.92 \pm 0.29$} \\ \cmidrule(l){2-14}
 &
  $\alpha$-loss &
  \multicolumn{2}{c}{$40.16 \pm 0.40$} &
  \multicolumn{2}{c}{$\mathbf{41.50 \pm 0.53}$} &
  \multicolumn{2}{c}{$\mathbf{43.01 \pm 0.69}$} &
  \multicolumn{2}{c}{$\mathbf{44.26 \pm 0.35}$} &
  \multicolumn{2}{c}{$\mathbf{45.95 \pm 0.22}$} &
  \multicolumn{2}{c}{$\mathbf{42.98 \pm 0.44}$} \\ \bottomrule
\end{tabular}}

\end{table*}

\begin{table*}[h!]
\refstepcounter{table}
\label{table:mce-asym10}
\centering
\renewcommand{\arraystretch}{0.9}
\resizebox{\textwidth}{!}{

\begin{tabular}{@{}ccclclclclclcl@{}}

\toprule
\multicolumn{2}{c}{Method} &
  \multicolumn{10}{c}{TABLE \ref{table:mce-asym10}: \textbf{Asymmetric} CIFAR-10} \\ \midrule
Augment &
  Loss &
  \multicolumn{2}{c}{$\eta$ = 0} &
  \multicolumn{2}{c}{0.1} &
  \multicolumn{2}{c}{0.2} &
  \multicolumn{2}{c}{0.3} &
  \multicolumn{2}{c}{0.4} &
  \multicolumn{2}{c}{Noisy Avg.} \\ \midrule
\multirow{4}{*}{NoAug} &
  CE &
  \multicolumn{2}{c}{$\mathbf{26.84 \pm 0.10}$} &
  \multicolumn{2}{c}{$30.19 \pm 0.31$} &
  \multicolumn{2}{c}{$32.77 \pm 0.51$} &
  \multicolumn{2}{c}{$35.32 \pm 0.11$} &
  \multicolumn{2}{c}{$38.42 \pm 0.45$} &
  \multicolumn{2}{c}{$34.18 \pm 0.35$} \\ \cmidrule(l){2-14} 
 &
  Focal &
  \multicolumn{2}{c}{$33.76 \pm 2.00$} &
  \multicolumn{2}{c}{$32.73 \pm 1.28$} &
  \multicolumn{2}{c}{$\mathbf{29.33 \pm 1.01}$} &
  \multicolumn{2}{c}{$\mathbf{31.48 \pm 0.27}$} &
  \multicolumn{2}{c}{$\mathbf{33.99 \pm 0.86}$} &
  \multicolumn{2}{c}{$\mathbf{31.88 \pm 0.86}$} \\ \cmidrule(l){2-14} 
 &
 
  NCE+RCE &
  \multicolumn{2}{c}{$30.02 \pm 0.30$} &
  \multicolumn{2}{c}{$29.68 \pm 0.70$} &
  \multicolumn{2}{c}{$30.57 \pm 0.51$} &
  \multicolumn{2}{c}{$32.01 \pm 1.06$} &
  \multicolumn{2}{c}{$36.89 \pm 0.20$} &
  \multicolumn{2}{c}{$32.29 \pm 0.62$} \\ \cmidrule(l){2-14}
 &
 
  $\alpha$-loss &
  \multicolumn{2}{c}{$28.98 \pm 0.85$} &
  \multicolumn{2}{c}{$\mathbf{28.79 \pm 0.21}$} &
  \multicolumn{2}{c}{$30.46 \pm 0.51$} &
  \multicolumn{2}{c}{$32.98 \pm 0.36$} &
  \multicolumn{2}{c}{$39.09 \pm 0.78$} &
  \multicolumn{2}{c}{$32.83 \pm 0.47$} \\ \midrule
  
\multirow{4}{*}{\textsc{AugMix}} &
  CE &
  \multicolumn{2}{c}{$12.67 \pm 0.38$} &
  \multicolumn{2}{c}{$13.42 \pm 0.36$} &
  \multicolumn{2}{c}{$14.97 \pm 0.37$} &
  \multicolumn{2}{c}{$16.63 \pm 0.57$} &
  \multicolumn{2}{c}{$20.27 \pm 0.70$} &
  \multicolumn{2}{c}{$16.32 \pm 0.50$} \\ \cmidrule(l){2-14} 
 &
  Focal &
  \multicolumn{2}{c}{$13.81 \pm 0.21$} &
  \multicolumn{2}{c}{$13.64 \pm 0.15$} &
  \multicolumn{2}{c}{$14.17 \pm 0.30$} &
  \multicolumn{2}{c}{$15.55 \pm 0.34$} &
  \multicolumn{2}{c}{$20.28 \pm 0.84$} &
  \multicolumn{2}{c}{$15.91 \pm 0.41$} \\ \cmidrule(l){2-14} 
 &

  NCE+RCE &
  \multicolumn{2}{c}{$13.13 \pm 0.15$} &
  \multicolumn{2}{c}{$13.45 \pm 0.11$} &
  \multicolumn{2}{c}{$13.86 \pm 0.17$} &
  \multicolumn{2}{c}{$\mathbf{14.74 \pm 0.06}$} &
  \multicolumn{2}{c}{$18.29 \pm 0.52$} &
  \multicolumn{2}{c}{$15.08 \pm 0.22$} \\ \cmidrule(l){2-14}
  
 &

  $\alpha$-loss &
  \multicolumn{2}{c}{$\mathbf{12.55 \pm 0.09}$} &
  \multicolumn{2}{c}{$\mathbf{12.84 \pm 0.2}$} &
  \multicolumn{2}{c}{$\mathbf{13.44 \pm 0.08}$} &
  \multicolumn{2}{c}{$14.88 \pm 0.33$} &
  \multicolumn{2}{c}{$\mathbf{17.95 \pm 0.36}$} &
  \multicolumn{2}{c}{$\mathbf{14.78 \pm 0.24}$} \\ \bottomrule

\end{tabular}}

\end{table*}

\begin{table*}[h!]
\refstepcounter{table}
\label{table:mce-asym100}
\centering
\renewcommand{\arraystretch}{0.9}
\resizebox{\textwidth}{!}{\begin{tabular}{@{}ccclclclclclcl@{}}
\toprule
\multicolumn{2}{c}{Method} &
  \multicolumn{10}{c}{TABLE \ref{table:mce-asym100}: \textbf{Asymmetric} CIFAR-100} \\ \midrule
Augment &
  Loss &
  \multicolumn{2}{c}{$\eta$ = 0} &
  \multicolumn{2}{c}{0.1} &
  \multicolumn{2}{c}{0.2} &
  \multicolumn{2}{c}{0.3} &
  \multicolumn{2}{c}{0.4} &
  \multicolumn{2}{c}{Noisy Avg.} \\ \midrule
\multirow{4}{*}{NoAug} &
  CE &
  \multicolumn{2}{c}{$53.92 \pm 0.06$} &
  \multicolumn{2}{c}{$59.20 \pm 0.32$} &
  \multicolumn{2}{c}{$63.11 \pm 0.09$} &
  \multicolumn{2}{c}{$66.58 \pm 0.38$} &
  \multicolumn{2}{c}{$69.98 \pm 0.22$} &
  \multicolumn{2}{c}{$64.72 \pm 0.25$} \\ \cmidrule(l){2-14} 
 &
  Focal &
  \multicolumn{2}{c}{$\mathbf{53.53 \pm 0.43}$} &
  \multicolumn{2}{c}{$57.15 \pm 0.40$} &
  \multicolumn{2}{c}{$60.28 \pm 0.09$} &
  \multicolumn{2}{c}{$63.70 \pm 0.67$} &
  \multicolumn{2}{c}{$67.64 \pm 0.36$} &
  \multicolumn{2}{c}{$62.19 \pm 0.38$} \\ \cmidrule(l){2-14} 
 &
 
  NCE+RCE &
  \multicolumn{2}{c}{$54.45 \pm 0.05$} &
  \multicolumn{2}{c}{$\mathbf{56.02 \pm 0.66}$} &
  \multicolumn{2}{c}{$\mathbf{57.19 \pm 0.26}$} &
  \multicolumn{2}{c}{$\mathbf{59.09 \pm 0.53}$} &
  \multicolumn{2}{c}{$\mathbf{61.34 \pm 0.26}$} &
  \multicolumn{2}{c}{$\mathbf{58.41 \pm 0.43}$} \\ \cmidrule(l){2-14}
 &
 
  $\alpha$-loss &
  \multicolumn{2}{c}{$54.58 \pm 0.12$} &
  \multicolumn{2}{c}{$56.07 \pm 0.32$} &
  \multicolumn{2}{c}{$58.59 \pm 0.25$} &
  \multicolumn{2}{c}{$61.41 \pm 0.31$} &
  \multicolumn{2}{c}{$65.25 \pm 0.26$} &
  \multicolumn{2}{c}{$60.33 \pm 0.28$} \\ \midrule
  
\multirow{4}{*}{\textsc{AugMix}} &
  CE &
  \multicolumn{2}{c}{$37.88 \pm 0.29$} &
  \multicolumn{2}{c}{$41.14 \pm 0.16$} &
  \multicolumn{2}{c}{$43.71 \pm 0.12$} &
  \multicolumn{2}{c}{$46.37 \pm 0.15$} &
  \multicolumn{2}{c}{$49.57 \pm 0.44$} &
  \multicolumn{2}{c}{$45.20 \pm 0.22$} \\ \cmidrule(l){2-14} 
 &
  Focal &
  \multicolumn{2}{c}{$\mathbf{36.11 \pm 0.20}$} &
  \multicolumn{2}{c}{$\mathbf{37.64 \pm 0.18}$} &
  \multicolumn{2}{c}{$\mathbf{39.77 \pm 0.10}$} &
  \multicolumn{2}{c}{$\mathbf{42.37 \pm 0.16}$} &
  \multicolumn{2}{c}{$\mathbf{45.52 \pm 0.27}$} &
  \multicolumn{2}{c}{$\mathbf{41.32 \pm 0.18}$} \\ \cmidrule(l){2-14} 
 &

  NCE+RCE &
  \multicolumn{2}{c}{$42.13 \pm 0.30$} &
  \multicolumn{2}{c}{$42.81 \pm 0.07$} &
  \multicolumn{2}{c}{$43.78 \pm 0.15$} &
  \multicolumn{2}{c}{$44.77 \pm 0.10$} &
  \multicolumn{2}{c}{$46.53 \pm 0.45$} &
  \multicolumn{2}{c}{$44.47 \pm 0.19$} \\ \cmidrule(l){2-14}
  
 &

  $\alpha$-loss &
  \multicolumn{2}{c}{$37.64 \pm 0.12$} &
  \multicolumn{2}{c}{$39.21 \pm 0.10$} &
  \multicolumn{2}{c}{$41.21 \pm 0.07$} &
  \multicolumn{2}{c}{$43.27 \pm 0.04$} &
  \multicolumn{2}{c}{$45.74 \pm 0.24$} &
  \multicolumn{2}{c}{$42.36 \pm 0.11$} \\ \bottomrule
  
\end{tabular}%
}

\caption*{TABLES I-V: \textbf{mCE (mean\% $\pm$ std)} over three random trials for the CIFAR-10, CIFAR-100, and Tiny ImageNet datasets corrupted by varied levels of \textit{synthetic} (symmetric or asymmetric) label noise. Each combination of dataset, noise rate, augmentation, and loss function is considered, and the average mCE across all nonzero noise rates for each method is reported in the Noisy Avg. column. The best result for each augmentation + dataset combination is \textbf{boldfaced}.}

\end{table*}

\begin{table*}[h]
\centering
\renewcommand{\arraystretch}{0.9}
\resizebox{\textwidth}{!}{\begin{tabular}{@{}ccclclclclclcl@{}}
\toprule
\multicolumn{2}{c}{Method} &
  \multicolumn{10}{c}{CIFAR-10N} &
  \multicolumn{2}{c}{CIFAR-100N} \\ \midrule
Augment &
  Loss &
  \multicolumn{2}{c}{Aggregate} &
  \multicolumn{2}{c}{Random 1} &
  \multicolumn{2}{c}{Random 2} &
  \multicolumn{2}{c}{Random 3} &
  \multicolumn{2}{c}{Worst} &
  \multicolumn{2}{c}{Noisy Fine} \\ \midrule
\multirow{4}{*}{NoAug} &
  CE &
  \multicolumn{2}{c}{$32.24 \pm 0.41$} &
  \multicolumn{2}{c}{$37.56 \pm 0.18$} &
  \multicolumn{2}{c}{$37.66 \pm 0.30$} &
  \multicolumn{2}{c}{$37.96 \pm 0.13$} &
  \multicolumn{2}{c}{$49.25 \pm 0.34$} &
  \multicolumn{2}{c}{$67.66 \pm 0.15$} \\ \cmidrule(l){2-14} 
 &
  Focal &
  \multicolumn{2}{c}{$29.85 \pm 0.42$} &
  \multicolumn{2}{c}{$34.84 \pm 0.46$} &
  \multicolumn{2}{c}{$34.85 \pm 0.52$} &
  \multicolumn{2}{c}{$35.20 \pm 0.39$} &
  \multicolumn{2}{c}{$48.05 \pm 0.96$} &
  \multicolumn{2}{c}{$66.13 \pm 0.18$} \\ \cmidrule(l){2-14} 
 &
 
  NCE+RCE &
  \multicolumn{2}{c}{$30.18 \pm 0.21$} &
  \multicolumn{2}{c}{$31.11 \pm 0.73$} &
  \multicolumn{2}{c}{$31.49 \pm 0.31$} &
  \multicolumn{2}{c}{$32.35 \pm 1.80$} &
  \multicolumn{2}{c}{$\mathbf{38.13 \pm 0.46}$} &
  \multicolumn{2}{c}{$\mathbf{62.82 \pm 0.34}$} \\ \cmidrule(l){2-14}
 &
 
  $\alpha$-loss &
  \multicolumn{2}{c}{$\mathbf{29.22 \pm 0.79}$} &
  \multicolumn{2}{c}{$\mathbf{30.71 \pm 1.18}$} &
  \multicolumn{2}{c}{$\mathbf{30.44 \pm 0.88}$} &
  \multicolumn{2}{c}{$\mathbf{31.34 \pm 0.36}$} &
  \multicolumn{2}{c}{$39.93 \pm 0.35$} &
  \multicolumn{2}{c}{$63.09 \pm 0.38$} \\ \midrule
  
\multirow{4}{*}{\textsc{AugMix}} &
  CE &
  \multicolumn{2}{c}{$15.40 \pm 0.30$} &
  \multicolumn{2}{c}{$18.59 \pm 0.15$} &
  \multicolumn{2}{c}{$18.76 \pm 0.19$} &
  \multicolumn{2}{c}{$18.95 \pm 0.17$} &
  \multicolumn{2}{c}{$29.73 \pm 0.28$} &
  \multicolumn{2}{c}{$52.52 \pm 0.32$} \\ \cmidrule(l){2-14} 
 &
  Focal &
  \multicolumn{2}{c}{$13.28 \pm 0.16$} &
  \multicolumn{2}{c}{$\mathbf{13.77 \pm 0.11}$} &
  \multicolumn{2}{c}{$\mathbf{13.60 \pm 0.30}$} &
  \multicolumn{2}{c}{$\mathbf{13.61 \pm 0.20}$} &
  \multicolumn{2}{c}{$24.31 \pm 0.18$} &
  \multicolumn{2}{c}{$49.47 \pm 0.18$} \\ \cmidrule(l){2-14} 
 &

  NCE+RCE &
  \multicolumn{2}{c}{$13.72 \pm 0.27$} &
  \multicolumn{2}{c}{$14.16 \pm 0.03$} &
  \multicolumn{2}{c}{$13.85 \pm 0.18$} &
  \multicolumn{2}{c}{$14.07 \pm 0.09$} &
  \multicolumn{2}{c}{$\mathbf{18.14 \pm 0.32}$} &
  \multicolumn{2}{c}{$48.90 \pm 0.05$} \\ \cmidrule(l){2-14}
  
 &

  $\alpha$-loss &
  \multicolumn{2}{c}{$\mathbf{13.06 \pm 0.13}$} &
  \multicolumn{2}{c}{$14.07 \pm 0.28$} &
  \multicolumn{2}{c}{$14.04 \pm 0.07$} &
  \multicolumn{2}{c}{$14.00 \pm 0.06$} &
  \multicolumn{2}{c}{$21.25 \pm 0.04$} &
  \multicolumn{2}{c}{$\mathbf{48.78 \pm 0.29}$} \\ \bottomrule
  
\end{tabular}}
\caption{\textbf{mCE (mean\% $\pm$ std)} over three random trials across CIFAR-*N for each combination of augmentation and loss function. \textit{Aggregate} through \textit{Worst} are corruptions of CIFAR-10 and \textit{Noisy Fine} is the sole corruption of CIFAR-100. The best result for each augmentation + dataset combination is \textbf{boldfaced}.}

\label{table:mce-human}

\end{table*}

\begin{figure*}[b!]
\centering
\parbox{5cm}{
\includegraphics[width=5.7cm]{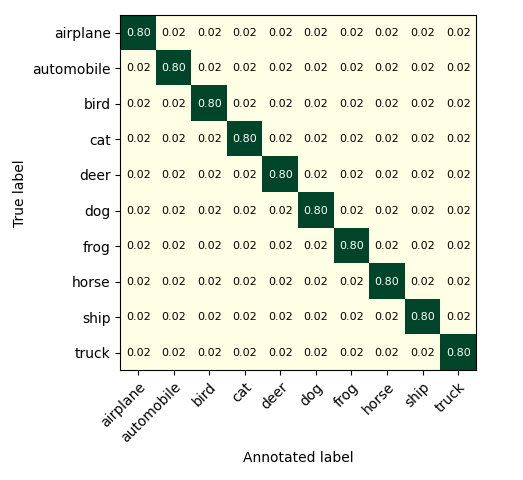}
\caption*{(a) Symmetric}
\label{fig:2figsA}}
\qquad
\begin{minipage}{5cm}
\includegraphics[width=5.7cm]{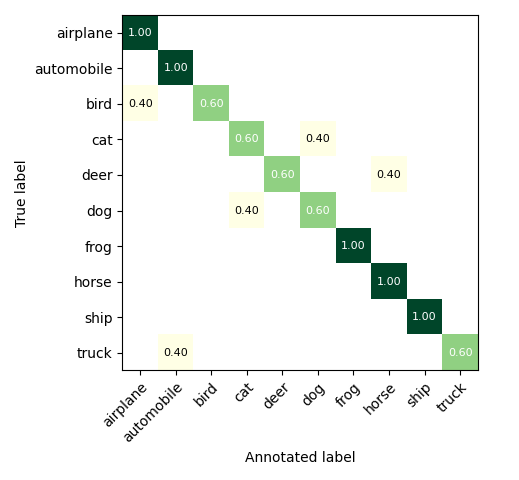}
\caption*{(b) Asymmetric}
\label{fig:2figsB}
\end{minipage}
\qquad
\begin{minipage}{5cm}
\includegraphics[width=5.7cm]{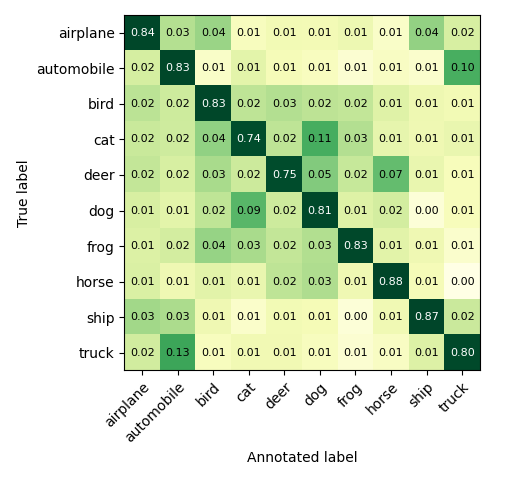}
\caption*{(c) CIFAR-10N Random 2}
\label{fig:2figsC}
\end{minipage}
\caption{Transition matrices for three methods of label noise generation, highlighting the greater diversity in human-annotated noisy labels (c) compared to its synthetically-generated counterparts (a, b).}
\label{fig:matrices}
\end{figure*}

\section{Experiments}

\label{section:experiments}

In the context of our novel task, we showcase \textsc{AugLoss} against previously-accepted SoTA methods that have been shown to combat label or feature corruption. In doing so, we explore how several realizations of \textsc{AugLoss} perform on corrupted datasets after training on varied settings of synthetic noise in the traditional datasets, or human-annotated noise in the CIFAR-*N datasets. 
In the next subsections, we detail the experimental protocol and provide brief justification for implementation choices.


\subsection{Datasets \& Metrics}

Previous efforts have used synthetic approaches to noise generation – namely symmetric and asymmetric noise – to evaluate model robustness under label corruption. In the symmetric case, the probability of a label flip is distributed evenly among the false classes: if \textit{A}, \textit{B}, \textit{C}, $\dots$, \textit{J} are classes and label \textit{A} were flipped to a different class, then \textit{B}, \textit{C}, $\dots$, \textit{J} would be chosen with equal probability. Symmetric noise is a consequence of class-\textit{independent} errors in the data-generating process – the same noise patterns are present no matter the difficulty in differentiating between classes. However, in crowdsourcing platforms (large-scale human-annotated labeling) inter-class visual similarity often influences the distribution of label noise in real-world datasets, giving rise to class-\textit{dependent} asymmetric noise as an alternative synthetic approach. 

Consistent with the standard, we train our models on the CIFAR-10, CIFAR-100, and Tiny ImageNet datasets corrupted by varied levels of symmetric (Tables \ref{table:mce-sym10}, \ref{table:mce-sym100}, \ref{table:mce-timgnet}) and asymmetric (Tables \ref{table:mce-asym10},\ref{table:mce-asym100}) noise. Each level is defined by the noise rate ($\eta$), which is the probability that any given label is flipped. Our choice of noise rates $ \eta \in \{ 0, 0.1, 0.2, 0.3, 0.4\}$ reflects the observation by ~\cite{pmlr-v97-song19b} that approximately $ 8\%$ to $38.5\%$ of labels in publicly-available image datasets are noisy. For asymmetric noise, we adhere to the standard mappings~\cite{ma2020normalized} outlined below:

\begin{itemize}
    \item \textbf{CIFAR-10:} some classes flip to other visually-similar ones – TRUCK $\mapsto$ AUTOMOBILE; BIRD $\mapsto$ AIRPLANE; DEER $ \mapsto$ HORSE; CAT $ \leftrightarrow$ DOG – and the rest simply flip to itself.
    \newline\newline\newline\newline\newline\newline\item \textbf{CIFAR-100:} we partition the label space of 100 classes into 20 superclasses, where each class belongs to a set containing four other visually-similar classes. Each class flips symmetrically to another member of its own superclass. 
\end{itemize}
Although synthetic approaches are commonplace in the evaluation of robust learning methods, we assert that human-annotated noise, when available, offers a more accurate representation of label corruption present in real-world systems. Specifically, we justify the choice of CIFAR-*N over synthetic noise with the following reasons~\cite{wei2021learning}:
\begin{enumerate}
    \item Datasets in CIFAR-*N provide more complex and diverse transition matrices compared to their synthetic counterparts, as shown in Figure~\ref{fig:matrices}. 
    \item Human-annotated noise is feature-dependent (\eg a cat-looking dog breed is more likely to be labeled a cat) while synthetic noise is strictly label-dependent.
    \item CIFAR-*N takes into account the potential co-existence of two classes in a single image, often found in the CIFAR-100 dataset.
\end{enumerate}
For these reasons, we additionally train our models on the CIFAR-*N datasets provided in Table \ref{table:mce-human}. The generation process of CIFAR-*N collects data from \textit{three} human annotators of CIFAR-10; therefore, each image in the dataset is annotated with \textit{three} independent labels. Considering the three labels reported for each CIFAR-10 image, (1) the \textbf{Aggregate} dataset selects the most common label, (2) the \textbf{Random 1/2/3} datasets uniformly sample the labels, and (3) the \textbf{Worst} dataset uniformly samples one of the incorrect labels (if they exist). On the other hand, CIFAR-*N only collects data from \textit{one} human annotator of CIFAR-100, and these results are directly reflected in the \textbf{Noisy Fine} dataset.

For similar reasons, we select CIFAR-*-C and Tiny ImageNet-C for evaluation. Since these datasets contains 15 corruptions of each image in their original dataset, we consider the \textit{mean corruption error} (mCE) – an average over the 15 individual corruption errors – as our primary performance metric. We consider the \textit{clean error} – the test error on clean datasets – as a baseline metric, and these results are reported in Tables \ref{table:clean-sym10}-\ref{table:clean-human} in Appendix \ref{section:cleanerrorresults}.

\subsection{Network Settings \& Preprocessing}


For CIFAR-*, we use a WideResNet-40-2 model \cite{zagoruyko2016wide}, and train for $100$ epochs; the optimizer is SGD with a Nesterov momentum of $0.9$ and weight decay of $5 \times 10^{-4}$; the learning rate scheduler is cosine annealing \cite{loshchilov2018cosine} with an initial value of $0.1$ and final value of $10^{-6}$. Additionally, train batches of size $32$ are preprocessed with random horizontal flips and batch normalization before being fed into the network.

For Tiny ImageNet, we use an EfficientNet-V2-M model \cite{mingxing2021efficient}, and train for $100$ epochs. The optimizer is SGD with a momentum of 0.9 and weight decay of $1 \times 10^{-4}$. We also use a cosine annealing learning rate scheduler with an initial learning rate of 0.05. Additionally, we apply random horizontal flips and batch normalization to batches of size 128 before being fed into the network.

\vspace{0.2in}

\begin{figure*}[t]
\centering
\includegraphics[width=18cm]{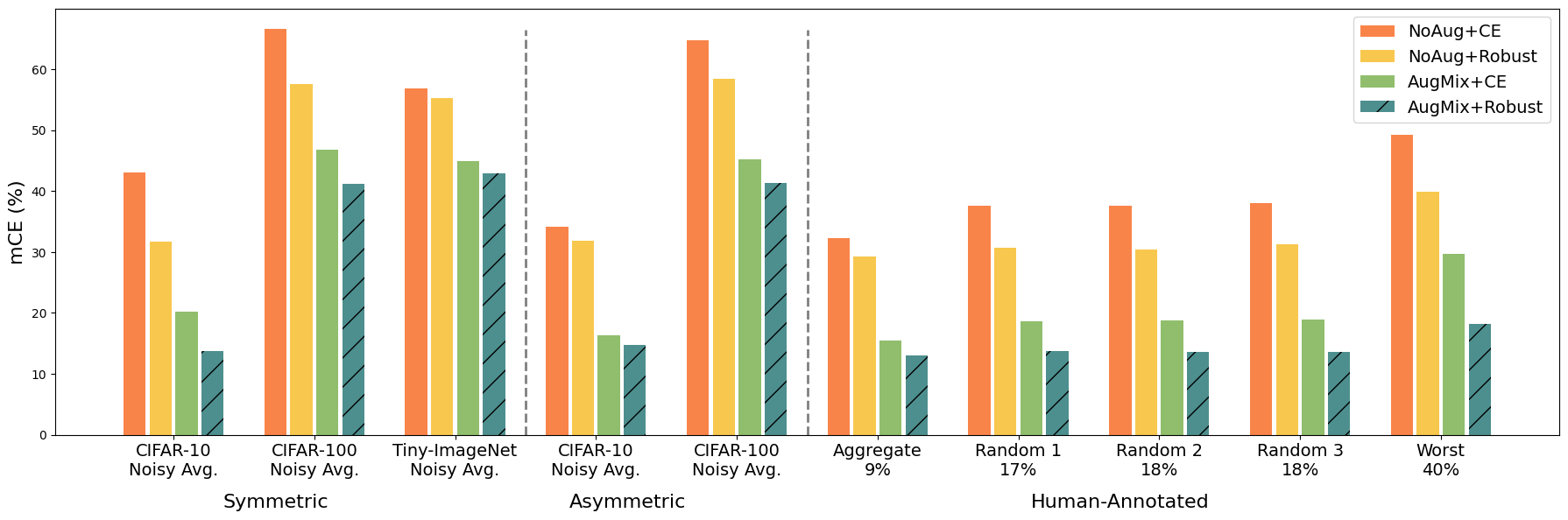}
\caption{The performances of each method type across symmetric, asymmetric, and human-annotated settings of label noise. The Noisy Avg. results for CIFAR-10, CIFAR-100, and Tiny ImageNet are included in the synthetic panels, while the CIFAR-10N results are included in the human-annotated panel. Hatched bars indicate the best performing method types for each setting. Note that our proposed methodology, \textsc{AugLoss}, \aka Augmix+Robust, is the clear winner in all settings considered.}
\label{fig:methodtypes}
\end{figure*}

\subsection{Augmentation + Loss Combinations}

We compare the performances of several $\textsc{AugLoss}$-specific methods in the settings outlined above. These methods are created by pairing known data augmentation techniques with a variety of basic loss functions. Specifically, we consider two types of data augmentation: \textbf{NoAug} (no augmentation) and the SoTA example \textbf{\textsc{AugMix}}, proposed by \cite{hendrycks2020augmix}. The \textsc{AugMix} technique is defined by its two nontrivial transformations, each composed of stochastically-sampled operations (\eg~autocontrast, rotate, solarize, \etc) organized in three chains of varied length, concluding with a mixture between each chain's output and the original image (see Figure \ref{fig:augmix}). Three generated images $x_{\text{orig}}, x_{\text{aug}1}, x_{\text{aug}2}$ are fed through the network to produce three posteriors $\hat{p}_{\text{orig}}, \hat{p}_{\text{aug}1}, \hat{p}_{\text{aug}2}$. The Jensen-Shannon divergence consistency loss ($\ell_{\text{JS}})$ serves as the augmentation regularizer \begin{equation}
\label{eq:js}
\ell_{\text{JS}}(\hat{P}) = \frac{1}{3} \sum\limits_{\hat{p}_{i} \in \hat{P}} \text{KL}\left( \hat{p}_{i} \| \hat{p}_{\text{mix}} \right), \end{equation} where $\hat{p}_{\text{mix}}$ is the mean of the three posteriors in $\hat{P} := \left( \hat{p}_{\text{orig}}, \hat{p}_{\text{aug}1}, \hat{p}_{\text{aug}2} \right)$. Consistent with \cite{hendrycks2020augmix}, $\lambda$ is set to 12.

For basic loss functions, we consider the standard \textbf{cross entropy} loss along with three tunable robust loss function families: \textbf{focal} loss, \textbf{NCE+RCE} loss, and $\boldsymbol\alpha$-loss. We tune on symmetric noise because: (1) this approach is computationally inexpensive and requires no extra information on class relationships; (2) we seek to evaluate each loss' ability to generalize its robust behavior to \textit{unforeseen} transition matrices and $\textit{varied}$ label noise rates.

\begin{figure*}[h]
\centering
\includegraphics[width=18cm]{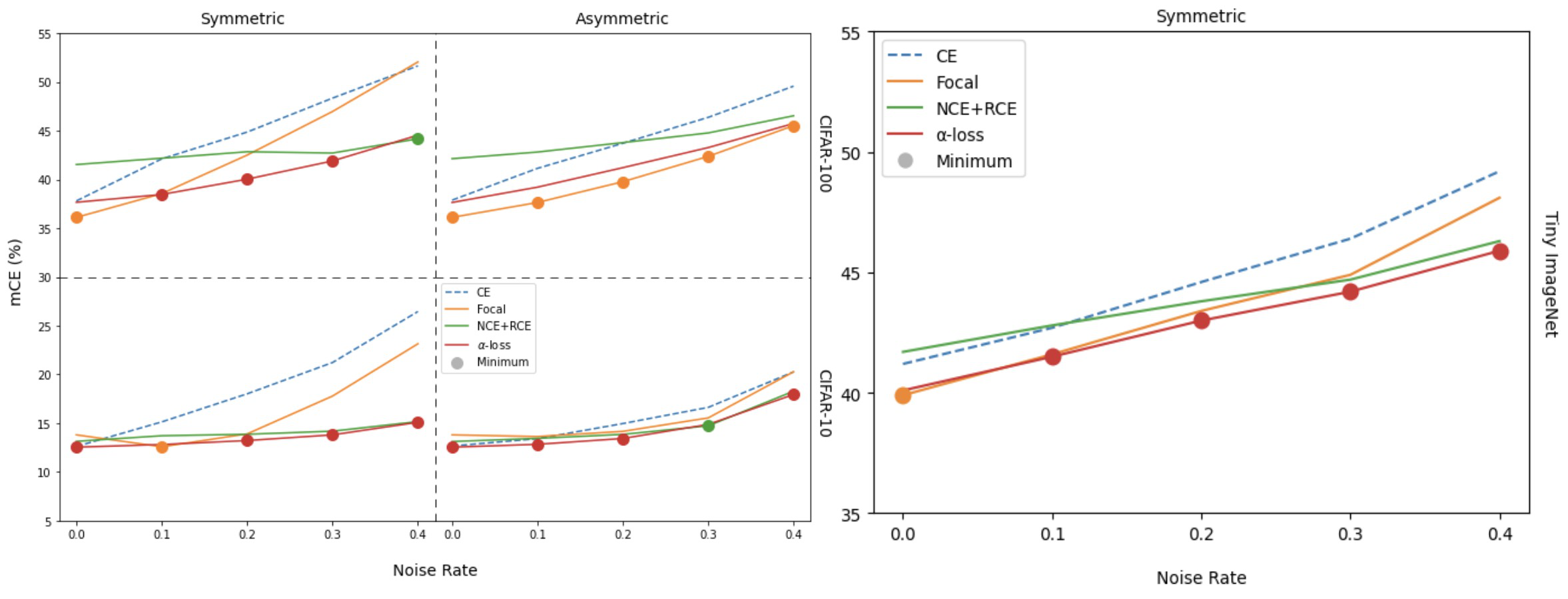}
\caption{The performances of each loss function across the synthetic (symmetric + asymmetric) settings of label noise. Dots are color-coded according to the best-performing loss functions at each setting and noise rate.}
\label{fig:lossgraph}
\end{figure*}

\subsection{Main Results \& Discussion}
For each pair of augmentation + loss combination and noisy-labeled train set, the mCE (mean\% $\pm$ standard deviation) is reported over three random trials in Tables \ref{table:mce-sym10}-\ref{table:mce-human}. Interpreting the results, two key observations are discussed below.

\subsection*{\textsc{AugLoss} best addresses dataset corruption.}

As previously stated, there exist widely-accepted methods that effectively combat label noise in the training stage (\ie robust loss without augmentation) or feature noise in the testing stage (\ie \textsc{AugMix}+CE paired with Jensen-Shannon loss), compared to the oft-used methodology, namely, CE loss without augmentation. Our experimental setup encapsulates these baseline methods to showcase the nontrivial gains achieved by \textsc{AugLoss} methods when training on noisy CIFAR-10, CIFAR-100  and testing on CIFAR-*-C, as well as when training on noisy Tiny Imagenet, and testing on Tiny Imagenet-C. Specifically, for analysis, we partition the augmentation + loss combinations into the following four method types:

\begin{itemize}
    \item \textbf{NoAug+CE}: the baseline method of CE loss without \textsc{AugMix}. These results are primarily included to highlight the need for a solution to combat both train-time label noise and test-time feature distribution shifts.
    \item \textbf{NoAug+Robust}: the group of widely-accepted methods that train with a robust loss function, specifically to combat label noise in the training stage. In this case, we group together three specific instances – NoAug+\textit{Focal}, NoAug+\textit{NCE+RCE}, and NoAug+\textit{$\alpha$-loss} – by averaging over the three results in each setting of train data.
    \item \textbf{\textsc{AugMix}+CE}: the specific method proposed by [21], using \textsc{AugMix} data augmentation with CE loss to combat data distribution shifts in the testing stage.
    \item \textbf{\textsc{AugMix}+Robust}: the group of methods that follow the \textsc{AugLoss} framework. Each method trains with \textsc{AugMix} and a robust loss function (focal, NCE+RCE, or $\alpha$-loss) in order to simultaneously harden the classifier against train-time label noise and test-time distribution shifts. As with \textit{NoAug+Robust}, we average over the results given by the three robust loss functions.
\end{itemize}
In Fig. \ref{fig:methodtypes}, we compare the performances (with respect to mCE) of each method type – NoAug+CE, NoAug+Robust, \textsc{AugMix}+CE, and \textsc{AugMix}+Robust –  under the different settings of label noise. Specifically, the two leftmost panels highlight the performance of each method type under synthetic (symmetric/asymmetric) label noise, while the rightmost panel considers the performance of each method type under human-annotated (CIFAR-10N) label noise. Taking both settings into account, the results show that the group of \textsc{AugLoss} methods (\textsc{AugMix}+Robust) consistently outperforms the other three types across all reported label noise settings. Although the well-known pairing of \textsc{AugMix} with CE loss already yields a drastic improvement in performance over the NoAug+CE baseline (\eg $-18.9\%$ mCE in the CIFAR-10N Random 2 setting), the novel pairing of \textsc{AugMix} with a \textit{robust} loss showcases the potential for further improvement in real-world noisy settings (\eg $-23.8\%$ mCE in the CIFAR-10N Random 2 setting). Overall, these results underscore the pressing need for models to contain \textit{both} data augmentation and robust loss functions (not just one) in order to simultaneously learn on noisy labels in the training stage, and generalize its classification ability to unforeseen feature distribution shifts in the testing stage.

\subsection*{No single robust loss function works best in all settings.}

Within the \textsc{AugLoss} framework, we seek to evaluate and compare the performances of each robust loss function family – focal, NCE+RCE, and $\alpha$-loss – paired with \textsc{AugMix} when training on noisy-labeled data and testing on CIFAR-*-C and Tiny ImageNet-C. In doing so, we observe that the best results in Tables 1-5 come from a \textit{mixture} of the three losses, not just one. This is illustrated amply in Fig. \ref{fig:lossgraph} where the performance of each loss function is plotted for varied settings of synthetic label noise. Here, we observe that $\alpha$-loss generally performs the best in both the symmetric label noise settings as well as the asymmetric CIFAR-10 setting. Additionally, NCE+RCE loss tends to perform worse in low levels of label noise – especially with CIFAR-100 and Tiny Imagenet – but shows to be competitive for relatively higher levels of label noise. In the three settings that $\alpha$-loss achieves the best results, focal loss appears to approach CE loss – the non-robust baseline – as label noise increases; however, focal loss clearly outperforms the others across all noise rates in the asymmetric CIFAR-100 setting, which underscores the idea that no robust loss function is the universal "best fit" for all settings of \textit{synthetic} label noise.

\begin{figure}[h]
\centering
\includegraphics[width=8cm]{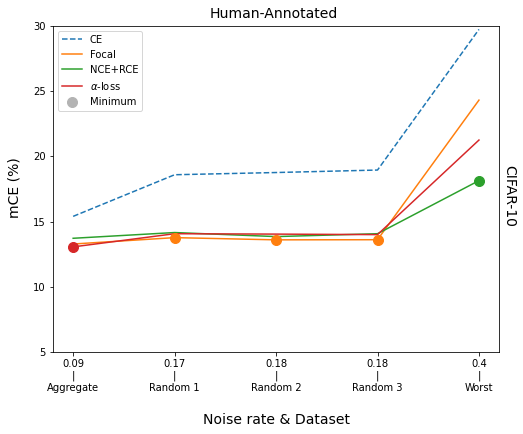}
\caption{The performances of each loss function across the five CIFAR-10N datasets. Dots are color-coded according to the best-performing loss functions for each dataset.}
\label{fig:lossgraphhuman}
\end{figure}

On the other hand, Figure \ref{fig:lossgraphhuman} displays the performance of each loss function in varied settings of human-annotated label noise– namely CIFAR-10N. Specifically, this figure shows that in our experiment, $\alpha$-loss performed the best with low-level noise (Aggregate), Focal loss performed the best with mid-level noise (Random 1-3), and NCE+RCE performed the best with higher levels of noise (Worst). Keeping in mind that each family is optimized under 20\% symmetric noise, these results suggest the following:

\begin{itemize}[leftmargin=*]
    \item $\alpha$-loss may be the appropriate choice in settings of label noise over-estimation, where $\alpha$ is tuned against a noise rate (\ie 20\%) higher than the true rate (\ie 9\%).
    \item Focal loss shows to be effective when the true noise rate is predictable (\ie 20\% $\approx$ 18\%).
    \item NCE+RCE loss appears to be valuable in settings of label noise under-estimation – a useful property in safety-critical applications – where its parameters are tuned against a lower noise level but can generalize to higher levels (\ie 40\%).
\end{itemize}
Since each loss clearly has its own merits, we conclude that no single choice appears to be the universal "best fit" within the \textsc{AugLoss} framework; rather, with more context on the task at hand, an appropriate loss function can be tailored to the specific needs of the problem.


\section{Conclusion}

\label{section:conclusion}

In this work, we have proposed a novel \textit{methodology}, \textsc{AugLoss}, wherein we have synthesized data augmentation techniques and robust loss functions in order to learn models that exhibit robustness to both train-time noisy labeling \textit{and} test-time feature distribution shifts.
In particular, using \textsc{AugMix} as the augmentation method, we have evaluated the performance of several tunable robust loss functions, including focal loss, NCE+RCE, and $\alpha$-loss.
Through our experimental procedure, we have demonstrated that on the whole, \textsc{AugLoss} yields much better mCE than either the robust loss functions or augmentation techniques alone. 
We have also observed that no tunable robust loss function stands out with different corruption settings handled better by different losses. 
Our novel methodology and the ensuing benchmarks significantly enhance existing results based on just loss functions or just augmentations. These benchmarks can further propel the field in identifying other augmentation techniques and robust loss functions to improve the robustness of models for out-of-distribution training and testing. In particular, a possible future effort is to leverage the recently introduced real-world WILDS dataset~\cite{koh2021wilds} to further evaluate the efficacy of \textsc{AugLoss}. Augmentations based on adversarial approaches \cite{wang2021augmax} and NoisyMix \cite{erichson2022noisymix} are other potential directions.

\section*{Acknowledgements}
The authors wish to thank Drs. Chaowei Xiao and Tyler Sypherd for engaging with us in discussions and sharing invaluable feedback with a preliminary version of this work. This work is supported in part by NSF grants CIF-1901243, CIF-1815361, CIF-2007688, CIF-2134256, and SaTC-2031799 as well as a Google AI for Social Good grant.


\section*{Additional Experiment Details}

\subsection*{Computing Resources}

Our experiments were executed on a cluster of K80 GPUs. The installed CUDA version was v11.2.
The settings with NoAug would generally take 3-4 hours to finish, while the settings with \textsc{AugMix} would generally take 8-10 hours.


\subsection*{Hyperparameter Tuning}

The focal loss is parameterized by $\gamma \in \mathbb{R}^{+}$ and we consider the following search space~\cite{focal2017}:
$$\gamma \in \{0.0,0.5,1.0,2.0,5.0\}.$$
Furthermore, the NCE+RCE loss employs two hyperparameters $(\beta_{1}, \beta_{2}) \in \mathbb{R}^{2}$ and motivated by~\cite{ma2020normalized}, we consider the following search space:
$$(\beta_{1}, \beta_{2}) \in \{0.1, 1.0, 10, 99, 99.9\} \times \{0.1, 1.0, 10, 100\}.$$
In the case of the Tiny ImageNet dataset, due to its increased complexity and variation in images, we extend this abovementioned search space for $\beta_{1}$ to accommodate higher values that might better capture the nuances of this dataset. Thus, the search space for the EfficientNet model trained on Tiny ImageNet dataset using the NCE+RCE loss is extended as follows:
\begin{align*}
(\beta_{1}, \beta_{2}) &\in \{0.1, 1.0, 10, 99, 99.9, 200, 500, 1000\} \\
&\quad \times \{0.1, 1.0, 10, 100\}.
\end{align*}
Lastly, the $\alpha$-loss is parameterized by $\alpha \in \mathbb{R}^{+}$ for which we consider the following search space~\cite{sypherd2022tunable}:
$$\alpha \in \{1.0,1.1,1.2,1.3,1.4,1.5,2.0,3.0,4.0\}$$
The best-performing parameters with respect to mCE for each family are shown below in Table \ref{table:hyperparameters}.

\setlength{\tabcolsep}{3pt}
\begin{table}[h!]
\centering
\begin{tabular}{@{}llllllllllll@{}}
\toprule
 & \multicolumn{3}{c}{Focal: $\gamma$} & \multicolumn{3}{c}{NCE+RCE: $(\beta_{1}, \beta_{2})$} & \multicolumn{3}{c}{$\alpha$-loss: $\alpha$} \\ \midrule
         & C10 & C100 & TIN & C10 & C100 & TIN & C10 & C100 & TIN    \\ \midrule
NA & 5.0 & 5.0 & 5.0 & (1.0,0.1) & (99.9,0.1) & (200,0.1) & 3.0 & 2.0 & 2.0 \\ \midrule
\textsc{AM} & 5.0 & 5.0 & 5.0 & (1.0,0.1) & (99,1.0) & (200,0.1) & 2.0 & 1.3 & 2.0   \\ \bottomrule
\end{tabular}
\caption{Hyperparameters for each tunable loss function. Abbreviations: C10 (CIFAR-10), C100 (CIFAR-100), TIN (Tiny ImageNet), NA (NoAug), AM (AugMix).}
\label{table:hyperparameters}
\end{table}


\bibliographystyle{IEEEtran}
\bibliography{IEEEabrv,ref.bib}

\onecolumn

\newpage

\appendices

\counterwithin{table}{section}
\counterwithin{figure}{section}

\section{Clean Error Results}

\label{section:cleanerrorresults}

\begin{table*}[h!]
\centering
\renewcommand{\arraystretch}{0.9}
\resizebox{\textwidth}{!}{\begin{tabular}{@{}ccclclclclclcl@{}}
\toprule
\multicolumn{2}{c}{Method} &
  \multicolumn{10}{c}{\textbf{Symmetric} CIFAR-10} \\ \midrule
Augment &
  Loss &
  \multicolumn{2}{c}{$\eta$ = 0} &
  \multicolumn{2}{c}{0.1} &
  \multicolumn{2}{c}{0.2} &
  \multicolumn{2}{c}{0.3} &
  \multicolumn{2}{c}{0.4} &
  \multicolumn{2}{c}{Noisy Avg.} \\ \midrule
\multirow{4}{*}{NoAug} &
  CE &
  \multicolumn{2}{c}{$\mathbf{5.36 \pm 0.21}$} &
\multicolumn{2}{c}{$11.18 \pm 0.23$} &
\multicolumn{2}{c}{$15.84 \pm 0.32$} &
\multicolumn{2}{c}{$21.25 \pm 0.30$} &
\multicolumn{2}{c}{$28.21 \pm 0.85$} &
\multicolumn{2}{c}{$19.12 \pm 0.42$} \\ \cmidrule(l){2-14}
 &
  Focal &
  \multicolumn{2}{c}{$10.30 \pm 1.79$} &
\multicolumn{2}{c}{$8.56 \pm 0.22$} &
\multicolumn{2}{c}{$13.54 \pm 0.21$} &
\multicolumn{2}{c}{$19.25 \pm 0.35$} &
\multicolumn{2}{c}{$26.74 \pm 0.49$} &
\multicolumn{2}{c}{$17.02 \pm 0.32$} \\ \cmidrule(l){2-14}
 &

  NCE+RCE &
  \multicolumn{2}{c}{$7.66 \pm 0.12$} &
\multicolumn{2}{c}{$8.13 \pm 0.26$} &
\multicolumn{2}{c}{$9.50 \pm 0.12$} &
\multicolumn{2}{c}{$10.53 \pm 0.13$} &
\multicolumn{2}{c}{$12.10 \pm 0.43$} &
\multicolumn{2}{c}{$10.06 \pm 0.23$} \\ \cmidrule(l){2-14}
 &

  $\alpha$-loss &
  \multicolumn{2}{c}{$6.15 \pm 0.08$} &
\multicolumn{2}{c}{$\mathbf{7.22 \pm 0.13}$} &
\multicolumn{2}{c}{$\mathbf{8.13 \pm 0.12}$} &
\multicolumn{2}{c}{$\mathbf{9.45 \pm 0.25}$} &
\multicolumn{2}{c}{$\mathbf{11.39 \pm 0.16}$} &
\multicolumn{2}{c}{$\mathbf{9.05 \pm 0.16}$} \\ \midrule

\multirow{4}{*}{\textsc{AugMix}} &
  CE &
  \multicolumn{2}{c}{$\mathbf{4.69 \pm 0.16}$} &
\multicolumn{2}{c}{$6.96 \pm 0.18$} &
\multicolumn{2}{c}{$9.70 \pm 0.05$} &
\multicolumn{2}{c}{$12.78 \pm 0.29$} &
\multicolumn{2}{c}{$18.32 \pm 0.71$} &
\multicolumn{2}{c}{$11.94 \pm 0.31$} \\ \cmidrule(l){2-14}
 &
  Focal &
  \multicolumn{2}{c}{$8.31 \pm 0.03$} &
\multicolumn{2}{c}{$6.46 \pm 0.14$} &
\multicolumn{2}{c}{$6.81 \pm 0.08$} &
\multicolumn{2}{c}{$9.88 \pm 0.43$} &
\multicolumn{2}{c}{$15.07 \pm 0.50$} &
\multicolumn{2}{c}{$9.55 \pm 0.29$} \\ \cmidrule(l){2-14}
 &

  NCE+RCE &
  \multicolumn{2}{c}{$6.25 \pm 0.04$} &
\multicolumn{2}{c}{$6.96 \pm 0.18$} &
\multicolumn{2}{c}{$7.33 \pm 0.11$} &
\multicolumn{2}{c}{$7.74 \pm 0.17$} &
\multicolumn{2}{c}{$8.83 \pm 0.21$} &
\multicolumn{2}{c}{$7.72 \pm 0.17$} \\ \cmidrule(l){2-14}

 &

  $\alpha$-loss &
  \multicolumn{2}{c}{$5.15 \pm 0.17$} &
\multicolumn{2}{c}{$\mathbf{5.70 \pm 0.21}$} &
\multicolumn{2}{c}{$\mathbf{6.08 \pm 0.20}$} &
\multicolumn{2}{c}{$\mathbf{7.07 \pm 0.18}$} &
\multicolumn{2}{c}{$\mathbf{8.40 \pm 0.24}$} &
\multicolumn{2}{c}{$\mathbf{6.81 \pm 0.21}$} \\ \bottomrule

\end{tabular}}

\caption{\textbf{Clean error (mean\% $\pm$ std)} over three random trials for the CIFAR-10 dataset corrupted by symmetric label noise. Each combination of dataset, noise rate, augmentation, and loss function is considered, and the average mCE across all nonzero noise rates for each method is reported in the Noisy Avg. column. The top result for each setting is \textbf{boldfaced}.}

\label{table:clean-sym10}

\end{table*}

\begin{table*}[h!]
\centering
\renewcommand{\arraystretch}{0.9}
\resizebox{\textwidth}{!}{\begin{tabular}{@{}ccclclclclclcl@{}}
\toprule
\multicolumn{2}{c}{Method} &
  \multicolumn{10}{c}{\textbf{Symmetric} CIFAR-100} \\ \midrule
Augment &
  Loss &
  \multicolumn{2}{c}{$\eta$ = 0} &
  \multicolumn{2}{c}{0.1} &
  \multicolumn{2}{c}{0.2} &
  \multicolumn{2}{c}{0.3} &
  \multicolumn{2}{c}{0.4} &
  \multicolumn{2}{c}{Noisy Avg.} \\ \midrule
\multirow{4}{*}{NoAug} &
  CE &
  \multicolumn{2}{c}{$\mathbf{24.67 \pm 0.39}$} &
\multicolumn{2}{c}{$32.01 \pm 0.14$} &
\multicolumn{2}{c}{$38.06 \pm 0.18$} &
\multicolumn{2}{c}{$44.08 \pm 0.32$} &
\multicolumn{2}{c}{$50.26 \pm 0.71$} &
\multicolumn{2}{c}{$41.1 \pm 0.34$} \\ \cmidrule(l){2-14}
 &
  Focal &
  \multicolumn{2}{c}{$25.8 \pm 0.55$} &
\multicolumn{2}{c}{$29.56 \pm 0.39$} &
\multicolumn{2}{c}{$35.16 \pm 0.39$} &
\multicolumn{2}{c}{$42.44 \pm 0.24$} &
\multicolumn{2}{c}{$49.7 \pm 0.12$} &
\multicolumn{2}{c}{$39.22 \pm 0.28$} \\ \cmidrule(l){2-14}
 &

  NCE+RCE &
  \multicolumn{2}{c}{$25.62 \pm 0.25$} &
\multicolumn{2}{c}{$\mathbf{26.84 \pm 0.4}$} &
\multicolumn{2}{c}{$29.27 \pm 0.35$} &
\multicolumn{2}{c}{$31.63 \pm 0.37$} &
\multicolumn{2}{c}{$35.33 \pm 0.08$} &
\multicolumn{2}{c}{$30.77 \pm 0.3$} \\ \cmidrule(l){2-14}
 &

  $\alpha$-loss &
  \multicolumn{2}{c}{$25.53 \pm 0.36$} &
\multicolumn{2}{c}{$27.08 \pm 0.4$} &
\multicolumn{2}{c}{$\mathbf{28.75 \pm 0.28}$} &
\multicolumn{2}{c}{$\mathbf{30.77 \pm 0.32}$} &
\multicolumn{2}{c}{$\mathbf{33.61 \pm 0.41}$} &
\multicolumn{2}{c}{$\mathbf{30.05 \pm 0.35}$} \\ \midrule

\multirow{4}{*}{\textsc{AugMix}} &
  CE &
  \multicolumn{2}{c}{$\mathbf{23.04 \pm 0.23}$} &
\multicolumn{2}{c}{$27.74 \pm 0.15$} &
\multicolumn{2}{c}{$31.18 \pm 0.61$} &
\multicolumn{2}{c}{$35.2 \pm 0.18$} &
\multicolumn{2}{c}{$39.34 \pm 0.17$} &
\multicolumn{2}{c}{$33.36 \pm 0.28$} \\ \cmidrule(l){2-14}
 &
  Focal &
  \multicolumn{2}{c}{$25.18 \pm 0.11$} &
\multicolumn{2}{c}{$26.08 \pm 0.36$} &
\multicolumn{2}{c}{$29.25 \pm 0.08$} &
\multicolumn{2}{c}{$34.03 \pm 0.22$} &
\multicolumn{2}{c}{$39.47 \pm 0.34$} &
\multicolumn{2}{c}{$32.21 \pm 0.25$} \\ \cmidrule(l){2-14}
 &

  NCE+RCE &
  \multicolumn{2}{c}{$26.5 \pm 0.67$} &
\multicolumn{2}{c}{$26.98 \pm 0.42$} &
\multicolumn{2}{c}{$28.23 \pm 0.2$} &
\multicolumn{2}{c}{$29.77 \pm 0.36$} &
\multicolumn{2}{c}{$31.43 \pm 0.23$} &
\multicolumn{2}{c}{$29.1 \pm 0.3$} \\ \cmidrule(l){2-14}

 &

  $\alpha$-loss &
  \multicolumn{2}{c}{$23.42 \pm 0.26$} &
\multicolumn{2}{c}{$\mathbf{24.9 \pm 0.04}$} &
\multicolumn{2}{c}{$\mathbf{26.92 \pm 0.27}$} &
\multicolumn{2}{c}{$\mathbf{29.41 \pm 0.32}$} &
\multicolumn{2}{c}{$\mathbf{32.43 \pm 0.52}$} &
\multicolumn{2}{c}{$\mathbf{28.42 \pm 0.29}$} \\ \bottomrule

\end{tabular}}

\caption{\textbf{Clean error (mean\% $\pm$ std)} over three random trials for the CIFAR-100 dataset corrupted by symmetric label noise. Each combination of dataset, noise rate, augmentation, and loss function is considered, and the average mCE across all nonzero noise rates for each method is reported in the Noisy Avg. column. The top result for each setting is \textbf{boldfaced}.}

\label{table:clean-sym100}

\end{table*}

\begin{table*}[h!]
\centering
\renewcommand{\arraystretch}{0.9}
\resizebox{\textwidth}{!}{

\begin{tabular}{@{}ccclclclclclcl@{}}

\toprule
\multicolumn{2}{c}{Method} &
  \multicolumn{10}{c}{\textbf{Asymmetric} CIFAR-10} \\ \midrule
Augment &
  Loss &
  \multicolumn{2}{c}{$\eta$ = 0} &
  \multicolumn{2}{c}{0.1} &
  \multicolumn{2}{c}{0.2} &
  \multicolumn{2}{c}{0.3} &
  \multicolumn{2}{c}{0.4} &
  \multicolumn{2}{c}{Noisy Avg.} \\ \midrule
\multirow{4}{*}{NoAug} &
  CE &
  \multicolumn{2}{c}{$\mathbf{5.21 \pm 0.26}$} &
\multicolumn{2}{c}{$8.4 \pm 0.45$} &
\multicolumn{2}{c}{$11.71 \pm 0.25$} &
\multicolumn{2}{c}{$15.71 \pm 0.53$} &
\multicolumn{2}{c}{$20.69 \pm 0.17$} &
\multicolumn{2}{c}{$14.13 \pm 0.35$} \\ \cmidrule(l){2-14}
 &
  Focal &
  \multicolumn{2}{c}{$9.83 \pm 2.05$} &
\multicolumn{2}{c}{$9.77 \pm 0.06$} &
\multicolumn{2}{c}{$\mathbf{8.4 \pm 0.11}$} &
\multicolumn{2}{c}{$\mathbf{10.56 \pm 0.18}$} &
\multicolumn{2}{c}{$\mathbf{13.77 \pm 0.27}$} &
\multicolumn{2}{c}{$\mathbf{10.62 \pm 0.16}$} \\ \cmidrule(l){2-14}
 &

  NCE+RCE &
  \multicolumn{2}{c}{$7.66 \pm 0.12$} &
\multicolumn{2}{c}{$8.22 \pm 0.34$} &
\multicolumn{2}{c}{$8.91 \pm 0.26$} &
\multicolumn{2}{c}{$11.16 \pm 0.35$} &
\multicolumn{2}{c}{$15.62 \pm 0.33$} &
\multicolumn{2}{c}{$10.98 \pm 0.32$} \\ \cmidrule(l){2-14}
 &

  $\alpha$-loss &
  \multicolumn{2}{c}{$6.34 \pm 0.13$} &
\multicolumn{2}{c}{$\mathbf{7.02 \pm 0.15}$} &
\multicolumn{2}{c}{$8.76 \pm 0.12$} &
\multicolumn{2}{c}{$12.29 \pm 0.31$} &
\multicolumn{2}{c}{$19.33 \pm 0.48$} &
\multicolumn{2}{c}{$11.85 \pm 0.26$} \\ \midrule

\multirow{4}{*}{\textsc{AugMix}} &
  CE &
  \multicolumn{2}{c}{$\mathbf{4.72 \pm 0.01}$} &
\multicolumn{2}{c}{$5.92 \pm 0.13$} &
\multicolumn{2}{c}{$7.31 \pm 0.3$} &
\multicolumn{2}{c}{$9.3 \pm 0.59$} &
\multicolumn{2}{c}{$13.01 \pm 0.09$} &
\multicolumn{2}{c}{$8.88 \pm 0.28$} \\ \cmidrule(l){2-14}
 &
  Focal &
  \multicolumn{2}{c}{$8.31 \pm 0.03$} &
\multicolumn{2}{c}{$8.33 \pm 0.08$} &
\multicolumn{2}{c}{$8.5 \pm 0.08$} &
\multicolumn{2}{c}{$9.71 \pm 0.28$} &
\multicolumn{2}{c}{$14.21 \pm 0.59$} &
\multicolumn{2}{c}{$10.19 \pm 0.26$} \\ \cmidrule(l){2-14}
 &

  NCE+RCE &
  \multicolumn{2}{c}{$6.25 \pm 0.04$} &
\multicolumn{2}{c}{$6.72 \pm 0.19$} &
\multicolumn{2}{c}{$7.27 \pm 0.08$} &
\multicolumn{2}{c}{$8.47 \pm 0.16$} &
\multicolumn{2}{c}{$12.14 \pm 0.47$} &
\multicolumn{2}{c}{$8.65 \pm 0.23$} \\ \cmidrule(l){2-14}

 &

  $\alpha$-loss &
  \multicolumn{2}{c}{$5.15 \pm 0.17$} &
\multicolumn{2}{c}{$\mathbf{5.6 \pm 0.17}$} &
\multicolumn{2}{c}{$\mathbf{6.44 \pm 0.05}$} &
\multicolumn{2}{c}{$\mathbf{8.04 \pm 0.3}$} &
\multicolumn{2}{c}{$\mathbf{11.54 \pm 0.42}$} &
\multicolumn{2}{c}{$\mathbf{7.91 \pm 0.23}$} \\ \bottomrule

\end{tabular}}

\caption{\textbf{Clean error (mean\% $\pm$ std)} over three random trials for the CIFAR-10 dataset corrupted by asymmetric label noise. Each combination of dataset, noise rate, augmentation, and loss function is considered, and the average mCE across all nonzero noise rates for each method is reported in the Noisy Avg. column. The top result for each setting is \textbf{boldfaced}.}

\label{table:clean-asym10}

\end{table*}


\begin{table*}[t!]
\renewcommand{\arraystretch}{0.9}
\centering
\resizebox{\textwidth}{!}{
\begin{tabular}{@{}ccclclclclclcl@{}}
\toprule
\multicolumn{2}{c}{Method} &
  \multicolumn{10}{c}{\textbf{Asymmetric} CIFAR-100} \\ \midrule
Augment &
  Loss &
  \multicolumn{2}{c}{$\eta$ = 0} &
  \multicolumn{2}{c}{0.1} &
  \multicolumn{2}{c}{0.2} &
  \multicolumn{2}{c}{0.3} &
  \multicolumn{2}{c}{0.4} &
  \multicolumn{2}{c}{Noisy Avg.} \\ \midrule
\multirow{4}{*}{NoAug} &
  CE &
  \multicolumn{2}{c}{$\mathbf{24.4 \pm 0.09}$} &
\multicolumn{2}{c}{$30.95 \pm 0.35$} &
\multicolumn{2}{c}{$36.83 \pm 0.19$} &
\multicolumn{2}{c}{$41.88 \pm 0.6$} &
\multicolumn{2}{c}{$48.08 \pm 0.67$} &
\multicolumn{2}{c}{$39.44 \pm 0.45$} \\ \cmidrule(l){2-14}
 &
  Focal &
  \multicolumn{2}{c}{$25.72 \pm 0.37$} &
\multicolumn{2}{c}{$29.25 \pm 0.29$} &
\multicolumn{2}{c}{$32.92 \pm 0.35$} &
\multicolumn{2}{c}{$38.35 \pm 0.23$} &
\multicolumn{2}{c}{$44.69 \pm 0.3$} &
\multicolumn{2}{c}{$36.3 \pm 0.29$} \\ \cmidrule(l){2-14}
 &

  NCE+RCE &
  \multicolumn{2}{c}{$25.45 \pm 0.14$} &
\multicolumn{2}{c}{$\mathbf{27.51 \pm 0.53}$} &
\multicolumn{2}{c}{$\mathbf{29.24 \pm 0.16}$} &
\multicolumn{2}{c}{$\mathbf{31.71 \pm 0.31}$} &
\multicolumn{2}{c}{$\mathbf{35.12 \pm 0.37}$} &
\multicolumn{2}{c}{$\mathbf{30.9 \pm 0.34}$} \\ \cmidrule(l){2-14}
 &

  $\alpha$-loss &
  \multicolumn{2}{c}{$25.66 \pm 0.24$} &
\multicolumn{2}{c}{$28.11 \pm 0.48$} &
\multicolumn{2}{c}{$30.91 \pm 0.35$} &
\multicolumn{2}{c}{$34.53 \pm 0.26$} &
\multicolumn{2}{c}{$40.45 \pm 0.26$} &
\multicolumn{2}{c}{$33.5 \pm 0.34$} \\ \midrule

\multirow{4}{*}{\textsc{AugMix}} &
  CE &
  \multicolumn{2}{c}{$\mathbf{23.28 \pm 0.45}$} &
\multicolumn{2}{c}{$26.76 \pm 0.29$} &
\multicolumn{2}{c}{$30.17 \pm 0.32$} &
\multicolumn{2}{c}{$33.02 \pm 0.48$} &
\multicolumn{2}{c}{$36.84 \pm 0.51$} &
\multicolumn{2}{c}{$31.7 \pm 0.4$} \\ \cmidrule(l){2-14}
 &
  Focal &
  \multicolumn{2}{c}{$25.18 \pm 0.11$} &
\multicolumn{2}{c}{$25.67 \pm 0.33$} &
\multicolumn{2}{c}{$\mathbf{27.18 \pm 0.35}$} &
\multicolumn{2}{c}{$\mathbf{29.54 \pm 0.18}$} &
\multicolumn{2}{c}{$\mathbf{32.85 \pm 0.18}$} &
\multicolumn{2}{c}{$\mathbf{28.81 \pm 0.26}$} \\ \cmidrule(l){2-14}
 &

  NCE+RCE &
  \multicolumn{2}{c}{$26.39 \pm 0.43$} &
\multicolumn{2}{c}{$27.31 \pm 0.54$} &
\multicolumn{2}{c}{$28.97 \pm 0.1$} &
\multicolumn{2}{c}{$30.82 \pm 0.34$} &
\multicolumn{2}{c}{$33.4 \pm 0.54$} &
\multicolumn{2}{c}{$30.12 \pm 0.38$} \\ \cmidrule(l){2-14}

 &

  $\alpha$-loss &
  \multicolumn{2}{c}{$23.38 \pm 0.26$} &
\multicolumn{2}{c}{$\mathbf{25.64 \pm 0.47}$} &
\multicolumn{2}{c}{$28.05 \pm 0.11$} &
\multicolumn{2}{c}{$30.35 \pm 0.31$} &
\multicolumn{2}{c}{$33.53 \pm 0.64$} &
\multicolumn{2}{c}{$29.39 \pm 0.38$} \\ \bottomrule

\end{tabular}
}
\caption{\textbf{Clean error (mean\% $\pm$ std)} over three random trials for the CIFAR-100 dataset corrupted by asymmetric label noise. Each combination of dataset, noise rate, augmentation, and loss function is considered, and the average mCE across all nonzero noise rates for each method is reported in the Noisy Avg. column. The top result for each setting is \textbf{boldfaced}.}
\label{table:clean-asym100}
\vspace{0.7cm}
\resizebox{\textwidth}{!}{
\begin{tabular}{@{}ccclclclclclcl@{}}
\toprule
\multicolumn{2}{c}{Method} &
  \multicolumn{10}{c}{CIFAR-10N} &
  \multicolumn{2}{c}{CIFAR-100N} \\ \midrule
Augment &
  Loss &
  \multicolumn{2}{c}{Aggregate} &
  \multicolumn{2}{c}{Random 1} &
  \multicolumn{2}{c}{Random 2} &
  \multicolumn{2}{c}{Random 3} &
  \multicolumn{2}{c}{Worst} &
  \multicolumn{2}{c}{Noisy Fine} \\ \midrule
\multirow{4}{*}{NoAug} &
  CE &
  \multicolumn{2}{c}{$10.96 \pm 0.29$} &
  \multicolumn{2}{c}{$16.09 \pm 0.20$} &
  \multicolumn{2}{c}{$16.14 \pm 0.36$} &
  \multicolumn{2}{c}{$15.98 \pm 0.13$} &
  \multicolumn{2}{c}{$30.60 \pm 0.27$} &
  \multicolumn{2}{c}{$47.21 \pm 0.29$} \\ \cmidrule(l){2-14} 
 &
  Focal &
  \multicolumn{2}{c}{$9.37 \pm 0.13$} &
  \multicolumn{2}{c}{$13.75 \pm 0.25$} &
  \multicolumn{2}{c}{$13.92 \pm 0.09$} &
  \multicolumn{2}{c}{$14.06 \pm 0.08$} &
  \multicolumn{2}{c}{$29.50 \pm 0.11$} &
  \multicolumn{2}{c}{$45.09 \pm 0.47$} \\ \cmidrule(l){2-14} 
 &
 
  NCE+RCE &
  \multicolumn{2}{c}{$9.11 \pm 0.20$} &
  \multicolumn{2}{c}{$9.86 \pm 0.11$} &
  \multicolumn{2}{c}{$10.34 \pm 0.26$} &
  \multicolumn{2}{c}{$10.21 \pm 0.28$} &
  \multicolumn{2}{c}{$\mathbf{17.13 \pm 0.39}$} &
  \multicolumn{2}{c}{$\mathbf{40.25 \pm 0.37}$} \\ \cmidrule(l){2-14}
 &
 
  $\alpha$-loss &
  \multicolumn{2}{c}{$\mathbf{8.23 \pm 0.26}$} &
  \multicolumn{2}{c}{$\mathbf{9.64 \pm 0.05}$} &
  \multicolumn{2}{c}{$\mathbf{9.94 \pm 0.25}$} &
  \multicolumn{2}{c}{$\mathbf{9.86 \pm 0.22}$} &
  \multicolumn{2}{c}{$19.36 \pm 0.38$} &
  \multicolumn{2}{c}{$40.94 \pm 0.33$} \\ \midrule
  
\multirow{4}{*}{\textsc{AugMix}} &
  CE &
  \multicolumn{2}{c}{$7.93 \pm 0.28$} &
  \multicolumn{2}{c}{$10.93 \pm 0.19$} &
  \multicolumn{2}{c}{$10.82 \pm 0.21$} &
  \multicolumn{2}{c}{$11.24 \pm 0.30$} &
  \multicolumn{2}{c}{$22.85 \pm 0.17$} &
  \multicolumn{2}{c}{$41.01 \pm 0.58$} \\ \cmidrule(l){2-14} 
 &
  Focal &
  \multicolumn{2}{c}{$8.00 \pm 0.23$} &
  \multicolumn{2}{c}{$7.93 \pm 0.15$} &
  \multicolumn{2}{c}{$7.62 \pm 0.25$} &
  \multicolumn{2}{c}{$7.93 \pm 0.04$} &
  \multicolumn{2}{c}{$17.85 \pm 0.16$} &
  \multicolumn{2}{c}{$38.80 \pm 0.27$} \\ \cmidrule(l){2-14} 
 &

  NCE+RCE &
  \multicolumn{2}{c}{$7.20 \pm 0.13$} &
  \multicolumn{2}{c}{$7.94 \pm 0.08$} &
  \multicolumn{2}{c}{$7.87 \pm 0.11$} &
  \multicolumn{2}{c}{$7.70 \pm 0.10$} &
  \multicolumn{2}{c}{$\mathbf{12.81 \pm 0.30}$} &
  \multicolumn{2}{c}{$38.10 \pm 0.06$} \\ \cmidrule(l){2-14}
  
 &

  $\alpha$-loss &
  \multicolumn{2}{c}{$\mathbf{6.29 \pm 0.05}$} &
  \multicolumn{2}{c}{$\mathbf{7.43 \pm 0.14}$} &
  \multicolumn{2}{c}{$\mathbf{7.53 \pm 0.12}$} &
  \multicolumn{2}{c}{$\mathbf{7.64 \pm 0.07}$} &
  \multicolumn{2}{c}{$15.88 \pm 0.39$} &
  \multicolumn{2}{c}{$\mathbf{38.07 \pm 0.08}$} \\ \bottomrule
  
\end{tabular}
}
\caption{\textbf{Clean error (mean\% $\pm$ std)} over three random trials across CIFAR-*N for each combination of augmentation and loss function. \textit{Aggregate} through \textit{Worst} are corruptions of CIFAR-10 and \textit{Noisy Fine} is the sole corruption of CIFAR-100. The best result for each augmentation + dataset combination is \textbf{boldfaced}.}
\label{table:clean-human}
\end{table*}

\newpage

\section{Example of AugMix}

\begin{figure*}[h!]
\centering
\includegraphics[width=16cm]{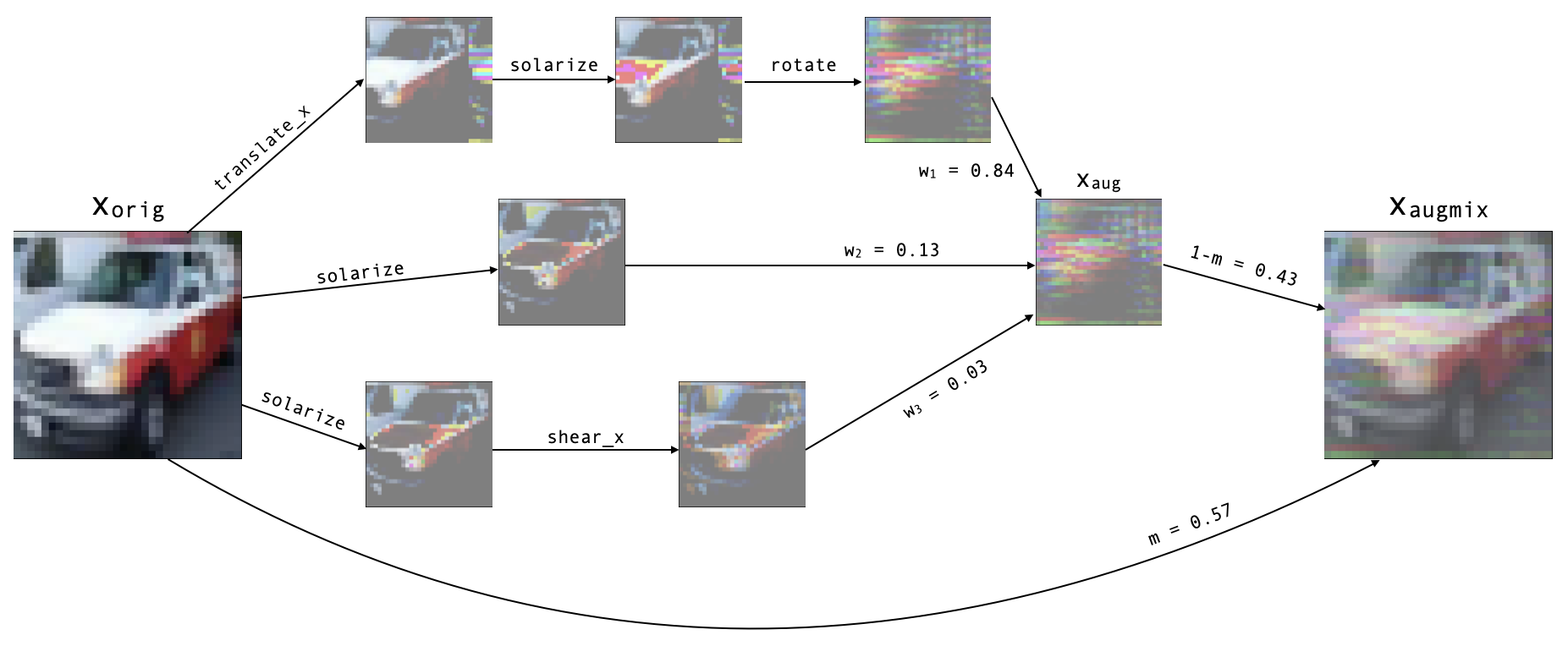}
\caption{Realization of \textsc{AugMix} on a TRUCK-labeled image.}
\label{fig:augmix}
\end{figure*}

In Figure \ref{fig:augmix}, we illustrate one realization of \textsc{AugMix}, highlighting the preservation of image semantics.

%

\end{document}